\documentclass[12pt, draftclsnofoot, onecolumn]{IEEEtran}
\usepackage{graphicx}
\usepackage{epstopdf}
\usepackage{amsthm}
\usepackage{epsfig}
\usepackage{latexsym}
\usepackage{amsfonts}
\usepackage{here}
\usepackage{rawfonts}
\usepackage[latin1]{inputenc}
\usepackage[T1]{fontenc}
\usepackage{calc}
\usepackage{capitalgreekitalic}
\usepackage{url}
\usepackage{enumerate}
\usepackage{color}
\usepackage[tbtags]{amsmath}
\usepackage{amssymb}
\usepackage{upref}
\usepackage{epic,eepic}
\usepackage{times}
\usepackage{dsfont}
\usepackage{comment}
\usepackage{cite}
\usepackage{bbm}
\usepackage{bm}
\usepackage{amsmath}

\allowdisplaybreaks[4]

\newtheorem{theorem}{\bf Theorem}

\newtheorem{lemma}{\bf Lemma}

\usepackage{dsfont}

\newcounter{step}
\newlength{\totlinewidth}
\newenvironment{algorithm}{%
  \rule{\linewidth}{1pt}
  \begin{list}{}%
    {\usecounter{step}%
      \settowidth{\labelwidth}{\textbf{Step 2:}}%
      \setlength{\leftmargin}{\labelwidth}%
      \setlength{\topsep}{-2pt}%
      \addtolength{\leftmargin}{\labelsep}%
      \addtolength{\leftmargin}{2mm}%
      \setlength{\rightmargin}{2mm}%
      \setlength{\totlinewidth}{\linewidth}%
      \addtolength{\totlinewidth}{\leftmargin}%
      \addtolength{\totlinewidth}{\rightmargin}%
      \setlength{\parsep}{0mm}%
      \raggedright}}%
  {\end{list}%
  \rule{\linewidth}{1pt}}
\newcounter{substep}

  {\end{list}}

\newlength{\aligntop}
\newlength{\alignbot}
 \makeatletter
\renewenvironment{align}{%
  \vspace{\aligntop}
  \start@align\@ne\st@rredfalse\m@ne
}{%
  \math@cr \black@\totwidth@
  \egroup
  \ifingather@
    \restorealignstate@
    \egroup
    \nonumber
    \ifnum0=`{\fi\iffalse}\fi
  \else
    $$%
  \fi
  \ignorespacesafterend%
  \vspace{\alignbot}\par\noindent
} \makeatother

\IEEEoverridecommandlockouts

\usepackage{algorithm}
\usepackage{algorithmic}

\makeatletter
\newcommand\semihuge{\@setfontsize\semihuge{19.3}{25}}
\makeatother

\makeatletter
\newcommand\semismall{\@setfontsize\semihuge{12.4}{15}}
\makeatother

\usepackage{subfigure}

\begin{document}
\title{\LARGE Performance Optimization for Variable Bitwidth Federated Learning in Wireless Networks}


\author{\normalsize{Sihua Wang,} \emph{Student Member, IEEE}, {Mingzhe Chen,} \emph{Member, IEEE},\\
	{Christopher G. Brinton}, \emph{Senior Member, IEEE}, {Changchuan Yin}, \emph{Senior Member, IEEE},  \\
	Walid Saad, \emph{Fellow, IEEE}, and {Shuguang Cui}, \emph{Fellow}, \emph{IEEE} \\
	\thanks{\scriptsize S. Wang and C. Yin are with the Beijing Laboratory of Advanced Information Network, and the Beijing Key Laboratory of Network System Architecture and Convergence, Beijing University of Posts and Telecommunications, Beijing 100876, China. Emails: \protect\url{sihuawang@bupt.edu.cn;} ccyin@ieee.org.}
	\thanks{\scriptsize M. Chen is with the Department of Electrical and Computer Engineering and Institute for Data Science and Computing, University of Miami, Coral Gables, FL, 33146 USA (Email: \protect\url{mingzhe.chen@miami.edu}).}
	\thanks{\scriptsize C. G. Brinton is with with the School of Electrical and Computer Engineering, Purdue University, West Lafayette, IN, USA, Email: \protect\url{cgb@purdue.edu}.}
	\thanks{\scriptsize W. Saad is with the Wireless@VT, Bradley Department of Electrical and Computer Engineering, Virginia Tech, Blacksburg, VA, 24060, USA, Email: \protect\url{walids@vt.edu}.}
	\thanks{\scriptsize S. Cui is currently with the School of Science and Engineering (SSE), the Future Network of Intelligence Institute (FNii), and the Guangdong Provincial Key Laboratory of Future Networks of Intelligence, the Chinese University of Hong Kong, and Shenzhen Research Institute of Big Data, Shenzhen, China, 518172; he is also affiliated with Peng Cheng Laboratory, Shenzhen, China, 518066, Email: \protect\url{shuguangcui@cuhk.edu.cn}.}
	\thanks{\scriptsize A preliminary version of this work \cite{NSGC} appears in the Proceedings of the 2022 IEEE Global Communications Conference (GLOBECOM).}
}

\maketitle

\pagestyle{empty}  
\thispagestyle{empty} 


\vspace{-1.5cm}

\begin{abstract}

This paper considers improving wireless communication and computation efficiency in federated learning (FL) via model quantization. In the proposed bitwidth FL scheme, edge devices train and transmit quantized versions of their local FL model parameters to a coordinating server, which, in turn, aggregates them into a quantized global model and synchronizes the devices. The goal is to jointly determine the bitwidths employed for local FL model quantization and the set of devices participating in FL training at each iteration. We pose this as an optimization problem that aims to minimize the training loss of quantized FL under a per-iteration device sampling budget and delay requirement. However, the formulated problem is difficult to solve without (i) a concrete understanding of how quantization impacts global ML performance and (ii) the ability of the server to construct estimates of this process efficiently. To address the first challenge, we analytically characterize how limited wireless resources and induced quantization errors affect the performance of the proposed FL method. Our results quantify how the improvement of FL training loss between two consecutive iterations depends on the device selection and quantization scheme as well as on several parameters inherent to the model being learned. Then, to address the second challenge, we show that the FL training process can be described as a Markov decision process (MDP) and  propose a model-based reinforcement learning (RL) method to optimize action selection over iterations. Compared to model-free RL, this model-based RL approach leverages the derived mathematical characterization of the FL training process to discover an effective device selection and quantization scheme without imposing additional device communication overhead. Simulation results show that the proposed FL algorithm can reduce the convergence time by 29\% and 63\% compared to a model free RL method and the standard FL method, respectively.

\end{abstract}



%
\IEEEpeerreviewmaketitle

\section{Introduction}

Federated learning (FL) is an emerging edge computing technology that enables a collection of devices to collaboratively train a shared machine learning model without sharing their collected data \cite{CGHS,YJJD,CJMK,LJSK,SJPYC}. During the FL training process, model parameters are trained locally on the device side and transmitted to a central center (e.g., at a base station (BS) coordinating the process across cellular devices) for global model aggregations. This procedure is repeated across several rounds until achieving an acceptable accuracy of the trained model \cite{DMPA,WZJD,ADSP,ZCWJC,JLD95}.

The local training and device-server communication processes can each have a significant impact on the performance of FL. These considerations are particularly important in resource-constrained edge settings in which devices exhibit heterogeneity in their communication and computation resources (e.g., a low-cost sensor vs. a high powered drone collecting measurements) \cite{hosseinalipour2020federated,kang2019incentive}. To minimize the resulting delays due to local training and parameter transmission, one promising method that has been recently proposed is the consideration of machine learning quantization at each device \cite{OYCS,KYJJ,DDCS,MJT}. In such schemes, the training and communication processes operate directly on quantized versions of the learning models, reducing the burden on device resources. However, efficient deployment of quantized FL over wireless networks poses several research challenges, related to the integration of quantization bitwidth considerations with the resulting FL training performance, which we study here.

\subsection{Related Works}
Recent works such as \cite{NMYH1,SCLY2,YYQT3,YSK4,JTGQ5,XT55,SCSYC55,Walid55,AJH,YYSM,THC,YSGD} have studied several important problems related to the implementation of quantized FL over wireless networks. The authors in \cite{NMYH1} designed a universal vector quantization scheme for FL model transmission to minimize the quantization error. In \cite{SCLY2}, a heterogeneous quantization framework was proposed for the FL model uploading process to speed up the convergence rate. A robust FL scheme was developed in \cite{YYQT3} to minimize the quantization errors and transmission outage probabilities under constraints on the training latency and device transmission powers. The authors in \cite{YSK4} proposed a hierarchical gradient quantization scheme for the FL framework to reduce the communication overhead while achieving similar learning performance. In \cite{JTGQ5}, the authors investigated a communication-efficient FL approach based on gradient quantization to alleviate the required communication bits and training rounds. \cite{XT55} further explored the impact of quantized communications on the performance of decentralized learning framework. The authors in \cite{SCSYC55} considered the extreme case of one-bit quantized local gradients for training the global FL model to reduce communication overhead. In \cite{Walid55}, the energy efficiency of a quantized FL scheme deployed over wireless networks is studied and the trade off between energy efficiency and accuracy is assessed. The authors in \cite{AJH} proposed an adaptive quantized gradient method to optimize the number of communication bits employed during the FL iterations so as to reduce communication energy. In \cite{YYSM}, an optimal vector quantizer was derived for minimizing the compression error of the local FL model update. In \cite{THC}, the authors proposed a quantized FL algorithm for a device-to-device based wireless system to reduce the data transmission volume of FL models between devices. The authors in \cite{YSGD} developed a methodology for jointly optimizing the loss function, cost for transmitting quantized FL models, and available wireless resources to reduce communication cost and training time.

These prior works on quantized FL have each assumed that certain key parameters of the model being learned -- such as smoothness and gradient diversity constants -- are known in advance of the training process. Under these assumptions, traditional optimization methods can be used to capture the relationship between quantization error and FL performance so as to find the optimal FL training policy. In practice, these model parameters cannot be obtained by the central server until the FL training process has completed, and thus, the solution derived by these traditional optimization methods may not be appropriate. To address this challenge, one promising approach is to employ reinforcement learning (RL) approaches \cite{MJLJ} for enabling the server to estimate these parameters over time through interaction with the devices during the training process, allowing discovery of a more effective FL policy.

Recently, a number of works \cite{WDWH6,PCCZ7,QQW8,HNJ9,YPS10,MXSX,YHL,YMFY} used RL algorithms to configure system parameters for FL performance optimization. In \cite{WDWH6}, the authors proposed a deep multi-agent RL to accelerate FL convergence while reducing the energy used for training. The authors in \cite{PCCZ7} designed a device selection scheme based on RL to minimize energy consumption and training delay, i.e., by searching for the most efficient set of devices to participate in each training iteration. A deep RL-based framework in \cite{QQW8} was proposed to maximize the long-term FL performance under energy and bandwidth constraints. In \cite{HNJ9}, the authors studied the use of a deep Q-network (DQN) to minimize wireless communication interruptions experienced by the FL framework due to device mobility. \cite{YPS10} used deep RL to jointly optimize training time and energy consumption via adjusting the CPU-cycle frequency of devices. The authors in \cite{MXSX} designed a DQN-based quantization allocation mechanism to improve the performance of FL. In \cite{YHL}, the authors employed a multiagent deep RL based quantization method to reduce the energy used for communication in FL framework. The authors in \cite{YMFY} analyzed the relationship between the global convergence and computational complexity in quantized a federated RL framework.

These prior works have thus employed RL methods to capture the relationship between FL performance and the training policy, in turn leading to improvements in different aspects of model training. However, with these methods, the coordinating server must collect numerous observations of different FL training policies by interacting with the devices over the environment, resulting in considerable delay for finding the optimal policy and encumbering FL convergence speed. To overcome this, we are motivated to develop \textit{model-based} RL methods based on mathematical models of the quantized FL training process. Specifically, the coordinating server will estimate the associated FL model parameters based on information captured during the training process, minimizing the time and overhead required to discover the optimal FL policy.
 
\subsection{Outline of Methodology and Contributions}

The main contribution of this paper is a novel methodology to optimize quantized FL algorithms over wireless networks by a model based RL method that can estimate the FL training parameters and mathematically model the FL training process without continual interacting with the devices. To our best knowledge, \emph{this is the first work that provides a systematic analysis of the integration of quantization bitwidth optimization into the FL framework.} Our key contributions include:


\begin{itemize}
	
    \item We propose a novel quantized FL framework in which distributed wireless devices train and transmit their locally trained FL models to a coordinating server based on variable bitwidths. The server selects an appropriate set of devices to execute the FL algorithm with variable quantized bitwidths in each iteration. To this end, we formulate the joint device selection and FL model quantization problem as an optimization problem whose goal is to minimize training loss while accounting for communication and computation heterogeneity as well as non-i.i.d. data distribution across the devices. We quantify these heterogeneity factors in terms of service delay and communication bandwidth requirements.
    
    \item To solve this problem, we first analytically characterize the expected training convergence rate of our quantized FL framework with non-i.i.d. data distribution. Our analysis shows how the expected improvement of FL training loss between two adjacent iterations depends on the device selection scheme, the quantization scheme, and inherent properties of the model being trained under the non-i.i.d. setting. To find the tightest bound, we introduce a linear regression method for estimating these model properties according to observable training information at the server. Given these estimates, we show that the FL training process can be mathematically described as a Markov decision process (MDP) with consecutive global model losses constituting state transitions.
         
    \item To learn the optimal solution of the formulated MDP, we construct a model-based RL method that infers the action (i.e., device selection and quantization scheme) which maximizes the expected reward (i.e., minimize global model loss) in each training iteration. Compared to traditional model-free RL approaches, our proposed method enables the server to optimize the FL training process through minimal interaction with each device. The removal of continual device-server communication requirements is particularly useful in the bandwidth-limited wireless edge settings we consider.
    
\end{itemize} 

Numerical evaluation results on real-world machine learning task datasets show that our proposed quantized FL methodology can reduce convergence time by up to 29\% while reducing the training iterations needed for convergence by up to 20\% compared with existing FL baselines. Additionally, these results show how the number of devices and number of quantization bits jointly affect the performance of FL over wireless networks.


\section{System Model and Problem Formulation}

\begin{figure}[t]
\centering
\setlength{\belowcaptionskip}{-0.45cm}
\vspace{-0.1cm}
\includegraphics[width=10cm]{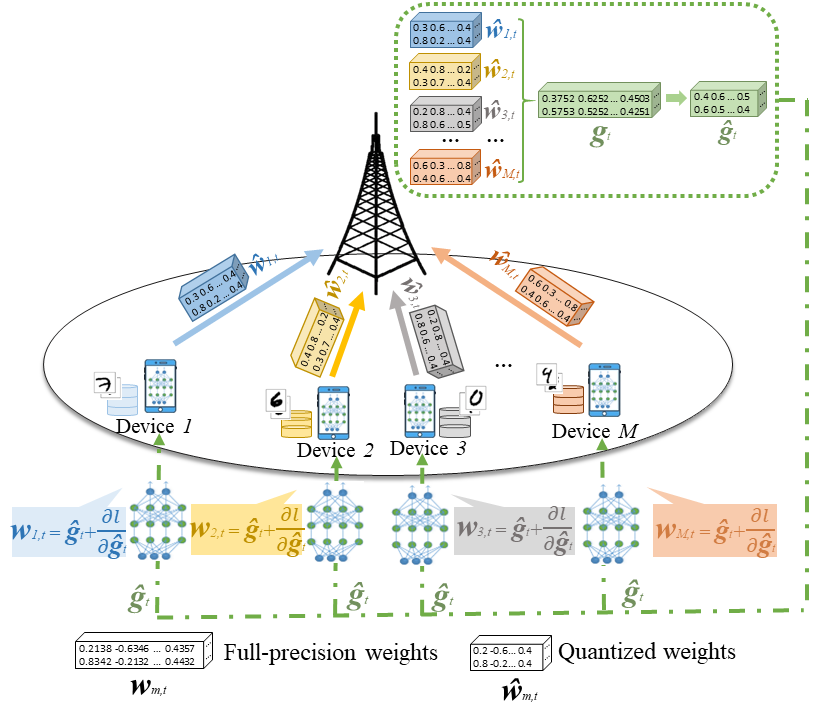}
\centering
\vspace{-0.5cm}
\caption{Depiction of our proposed low bitwidth federated learning methodology deployed over multiple devices and one base station in a wireless network.}
\vspace{-0.4cm}
\label{fig2}
\end{figure}

Consider a wireless network that consists of a set $\mathcal{M}$ of $M$ devices connected upstream to a coordinating server, which we will assume is a server without loss of generality. These devices are aiming to execute an FL algorithm for training a machine learning model, as shown in Fig. 1. Each device $m$ has $N_m$ training data samples, and each training data sample $n$ consists of an input feature vector $\bm x_{m,n} \in \mathbb{R}^{N_{I}\times 1}$ and (in the case of supervised learning) a corresponding label vector $\bm y_{m,n} \in \mathbb{R}^{N_{O}\times 1}$. The objective of the server and the devices is to minimize the global loss function over all data samples, which is given by
\begin{equation}\label{eq:FLloss}
\begin{aligned}
F(\bm g)=\mathop {\min }\limits_{\bm g} \frac{1}{N}\sum\limits_{m = 1}^M \sum\limits_{n = 1}^{N_m} f\left(\bm g,\bm x_{m,n},\bm y_{m,n}\right),
\end{aligned}
\end{equation}
where $\bm g \in \mathbb{R}^{Y \times 1}$ is a vector that captures the global FL model of dimension $Y$ trained across the devices, with $N=\sum\limits_{m = 1}^{M} N_m$ being the total number of training data samples of all devices. $f\left(\bm g,\bm x_{m,n},\bm y_{m,n}\right)$ is a loss function (e.g., squared error) that measures the accuracy of the generated global FL model $\bm g$ in building a relationship between the input vector $\bm x_{m,n}$ and the output vector $\bm y_{m,n}$.

\subsection{Training Process of Low Bitwidth Federated Learning}
In FL, devices and the server iteratively exchange their model parameters to find the optimal global model $\bm g$ that minimizes the global loss function in (\ref{eq:FLloss}). However, due to limited computational and wireless resources, devices may not be able to train and transmit such large sized model parameters (e.g., as in the case of deep learning). To reduce the computation and transmission delays, bitwidth federated learning was proposed in \cite{LWZE}. Compared to the widely studied case of federated averaging \cite{MZWC}, the FL model parameters in bitwidth FL are quantized. The overall training process of bitwidth FL is given as follows:
\begin{enumerate}
\item The server quantizes the initialized global learning model and broadcasts it to each device.

\item Each device calculates the training loss using the quantized global learning model and its collected data samples.

\item Based on the calculated training loss, the quantized~learning model in each device is updated.

\item Each device quantizes its updated learning model.

\item The server selects a subset of devices for local FL model transmission.

\item The server aggregates the collected local FL models into a global FL model that will be transmit to devices.
\end{enumerate}
Steps 2-6 are repeated until the optimal vector $\bm g$ is found.

From the training process, we see that, in bitwidth FL, each device uses a quantized FL model to calculate the training loss and gradient vectors during the training process. Therefore, the quantization scheme in bitwidth FL will affect the resource requirements of FL model training and transmission. This is significantly different from quantization-based FL algorithms \cite{LWZE} that must recover the quantized FL model during the training process, thus introducing additional computational complexity and reducing training efficiency. Next, we will introduce the training process mathematically.

\subsubsection{Calculation of Training Loss of Each Device}
We first introduce the calculation of each device's training loss for step 2. Without loss of generality, we will assume that a neural network is being trained; the quantization method can be used in other machine learning approaches (such as support vector machines (SVM) \cite{SVM}) as well.

The weights of each device's local FL model are quantized into $\alpha_{t}$ bits. Through this, the full-precision neural network is transformed into a quantized neural network (QNN). When $\alpha_{t}\!=\!1$, each QNN weight has two possible values, namely -1/0 or +1. Therefore, a neural network that consists of the weights with two possible values is called a binary neural network (BNN) \cite{BNN}. Given the input vector ${\bm h}^k_{m,t}$ and the weight vector $\hat{\bm g}^k_{t}$ of the neurons in layer $k$ that is represented by $\alpha_{t}$ bits, the output of each layer $k$ at iteration $t$ is given by \cite{SWDB}
\begin{equation}\label{eq:QNNForward}
\begin{aligned}
\bm h^{k+1}_{m,t}=\left\{ \begin{array}{l}
\sigma \left( {\bm h}^k_{m,t} \odot \hat{\bm g}^k_{t}\right), \;\;\; \! \qquad \qquad {\rm if} \;\; \alpha_{t}=1,\\
\sigma  \left( \sum\limits_{i = 0}^{\alpha_{t}-1} \sum\limits_{j = 0}^{\alpha_{t}-1}2^{i+j} ( { h}^{k,i}_{m,t} \odot \hat{ g}^{k,j}_{t})\right), \quad \!{\rm if} \;\; \alpha_{t}>1,\\
\end{array} \right.\
\end{aligned}
\end{equation}
where $\sigma(\cdot)$ is the activation function and $\odot$ represents the inner product for vectors with bitwise operations. Given the outputs of all neuron layers ${\bm h}_{m,t}=[{\bm h}^1_{m,t},\ldots,{\bm h}^K_{m,t}]$, the cross-entropy loss function can be expressed based on the neurons in an output layer ${\bm h}^K_{m,t}$ as
\begin{equation}\label{eq:QNNloss}
\begin{aligned}
f\left(\hat{\bm g}_{t},\bm x_{m,n},\bm y_{m,n}\right)= -\bm y_{m,n}{\rm log}({\bm h}^K_{m,t})+(1-\bm y_{m,n}){\rm log}(1-{\bm h}^K_{m,t}),
\end{aligned}
\end{equation}
where $\hat{\bm g}_{t}=[\hat{\bm g}^{1}_{t},...,\hat{\bm g}^{k}_{t},...,\hat{\bm g}^{K}_{t}]$ is the quantized global FL model.

\subsubsection{FL Model Update}
A backward propagation (BP) algorithm based on stochastic gradient descent is used to update the parameters in QNN. The update function is expressed as
\begin{equation}\label{eq:update1}
\begin{aligned}
\bm w_{m,t+1}&=\hat{\bm g}_{t}-\lambda \!\!\! \sum\limits_{n\in \mathcal{N}_{m,t}}\!\!\frac{\partial f\left(\bm g,\bm x_{m,n},\bm y_{m,n}\right)}{\bm \partial \bm g},
\end{aligned}
\end{equation}
where $\lambda$ is the learning rate, $\mathcal{N}_{m,t}$ is the subset of training data samples (i.e., minibatch) selected from device $m$'s training dataset $\mathcal{N}_m$ at iteration $t$, $\bm w_{m,t+1}$ is the updated local FL model of device $m$ at iteration $t\!+\!1$, and
\begin{equation}\label{eq:update2}
\begin{aligned}
\frac{\partial f}{\partial \bm g}\!=\!\frac{\partial f_{m,t}}{\partial \hat {\bm g}_{t}} \times \frac{\partial \hat {\bm g}_{t}}{\partial \bm g_{t}}\!=\!\frac{\partial f_{m,t}}{\partial \hat {\bm g}_{t}} \times {\rm Htanh}(\bm g_{t}),
\end{aligned}
\end{equation}
where $\bm g_{t}$ represents the full-precision weights. ${\rm Htanh}(x)\!=\!{\rm max}(-1,{\rm min}(1,x))$ is used to approximate the derivative of the quantization function that is not differentiable. From (\ref{eq:update1}) and (\ref{eq:update2}), we can see that the weights are updated with full-precision values since the changes of the learning model update at each step are small.

\subsubsection{FL Model Quantization at Device}
As each local FL model is updated, these full-precision weights must be completely quantized into $\alpha_{t}$ bits, which is given by\cite{CY}
\begin{equation}\label{eq:BBNN}
\begin{aligned}
\hat{ w}^{k,j}_{m,t}({\alpha_{t}})=Q({ w}^{k,j}_{m,t},\alpha_{t})=\left\{\begin{array}{l}
\;\;{\rm sign}({ w}^{k,j}_{m,t}), \quad \quad {\rm if} \; \alpha_{t} \!=\!1,\\
\!\!\!\frac{R\left((2^{\alpha_{t}}-1){w}^{k,j}_{m,t}\right)}{2^{\alpha_{t}}-1}, \;\; {\rm if} \; 1 \!<\! \alpha_{t} \!<\! V,\\
\;\;\;\;\;\;{ w}^{k,j}_{m,t},  \qquad \quad \;\; {\rm if} \;\alpha_{t} \!=\! V,\\
\end{array} \right.\
\end{aligned}
\end{equation}
where $V$ is the bitwidth of the full-precision and ${\rm sign}(x)=1$ if $x \geqslant 0$ and ${\rm sign}(x)=-1$, otherwise. $R(\cdot)$ is a rounding function with $R(x)=\lfloor x \rfloor$ if $x \leqslant \frac{\lfloor x \rfloor+\lceil x \rceil}{2}$, and $R(x)=\lceil x \rceil$, otherwise. From (\ref{eq:BBNN}), we see that when $\alpha_{t} =1$ (i.e., the binary case), if ${w}^{k,j}_{m,t} >0$, we have  $\hat{w}^{k,j}_{m,t}=1$ with ${w}^{k,j}_{m,t}$ and $\hat{w}^{k,j}_{m,t}$ being $j$-th element in ${w}^{k,j}_{m,t}$ and $\hat{w}^{k,j}_{m,t}$ otherwise $\hat{ w}^{k,j}_{m,t}=-1$. For $1 < \alpha_{t} < V$,  ${\bm w}^{k}_{m,t}$ is quantized with increasing precision between $-1$ and $1$. Finally, when $\alpha_t = V$, there is no quantization.

\subsubsection{FL Model Transmission and Aggregation}
Due to limited wireless bandwidth, the server may need to select a subset of devices to upload their local FL models for aggregation into the global model. Given the quantized local FL model $\hat{\bm w}_{m,t}$ of each device $m$ at each iteration $t$, the update of the global FL model at iteration $t$ is given by
\begin{equation}\label{eq:localloss}
\begin{aligned}
\bm g_t(\bm u_{t},{\alpha_{t}})=\sum\limits_{m = 1}^M \frac{u_{m,t}N_{m,t}}{\sum\limits_{m = 1}^M u_{m,t}N_{m,t}}  \hat{\bm w}_{m,t}({\alpha_{t}}),
\end{aligned}
\end{equation}
where $\frac{u_{m,t}N_{m,t}}{\sum\limits_{m = 1}^M u_{m,t}N_{m,t}}$ is a scaling update weight of $\hat{\bm w}_{m,t}$, with $N_{m,t}$ being the number of data samples used to train $\hat{\bm w}_{m,t}$ at device $m$. $\bm g_t(\bm u_t)$ is the global FL model at iteration $t$, and $\bm u_{t}=[u_{1,t},\ldots,u_{M,t}]$ is the device selection vector, with $u_{m,t}=1$ indicating that device $m$ will upload its quantized local FL model $\hat{\bm w}_{m,t}$ to the server at iteration $t$, and $u_{m,t}=0$ otherwise.

\subsubsection{FL Model Quantization at the server}
As the global FL model is aggregated based on the collected local FL models, the server must quantize it in low bitwidth that can be directly used to calculate the training loss at each device. This is given by \cite{YL}
\begin{equation}\label{eq:BBNN1}
\begin{aligned}
\hat{\bm g}^{k}_{t}=Q({\bm g^{k}_{t}},\alpha_{t})=\left\{\begin{array}{l}
\;\;{\rm sign}(\bm g^{k}_{t}), \quad \quad {\rm if} \; \alpha_{t} \!=\!1,\\
\!\!\!\frac{R\left((2^{\alpha_{t}}-1)\bm g^{k}_{t}\right)}{2^{\alpha_{t}}-1}, \;\; {\rm if} \;\; 1 \!<\! \alpha_{t} \!<\! V,\\
\;\;\;\;\;\;\bm g^{k}_{t},  \qquad \quad {\rm if} \;\;\alpha_{t} \!=\! V.\\
\end{array} \right.\
\end{aligned}
\end{equation}

\subsection{Training Delay of Low Bitwidth Federated Learning}
We next study the training delay of bitwidth FL. From the training steps, we can see that the delay consists of four components: (a) time used to calculate the training loss, (b) FL model update delay, (c) FL model quantization delay, and (d) FL model transmission delay. However, the FL model update delay is unrelated to the number of quantization bits $\alpha_{t}$, since the models are updated with full-precision values. Thus, component (b) is constant with respect to our methodology and can be ignored. Then, the training delay is specified as follows:

\subsubsection{Time Used to Calculate the Training Loss}
The time used to calculate the training loss depends on the number of multiplication operations in (\ref{eq:QNNForward}) and (\ref{eq:QNNloss}). From (\ref{eq:QNNForward}), we can see that the computational complexity of each multiplication operation is related to the number of bits $\alpha_{t}$ used to represent each element in FL model vector. Specifically, given $\alpha_{t}$, the time used to calculate the training loss is given by
\begin{equation}\label{eq:QNNdelay}
\begin{aligned}
l^{\rm C}_{m,t}(\alpha_{t})=\rho\frac{\alpha_{t}^2N^{\rm C}}{\vartheta f},
\end{aligned}
\end{equation}
where $\rho$ is the time consumption coefficient depending on the chip of each device and $N^{\rm C}$ is the number of multiplication operations in the neural network. $f$ and $\vartheta$ represent the frequency of the central processing unit (CPU) and the number of bits that can be processed by the CPU in one clock cycle, respectively.


\subsubsection{FL Model Quantization Delay}
Since the updated local FL model is in full-precision, each device must quantize its updated local FL model using (\ref{eq:BBNN}) to reduce transmission delay. Given $\alpha_{t}$, the quantization delay can be represented as \cite{SMXC}
\begin{equation}\label{eq:round1}
\begin{aligned}
l^{\rm Q}_{m,t}(\alpha_{t})=\left\{ \begin{array}{l}
0, \quad {\rm if} \; \alpha_{t} = 1 \;{\rm or}\; \alpha_{t} = V,\\
\frac{D}{\vartheta f},  \;\;\;\;{\rm if} \;\;  1 < \alpha_{t} < V,\\
\end{array} \right.\
\end{aligned}
\end{equation}
where $D$ is the number of neurons in the neural network. In (\ref{eq:round1}), when $\alpha_{t}\!=\!1$ or $\alpha_{t}\!=\!V$, the quantization delay will be 0. When $\alpha_{t}\!=\!1$, the value of quantized weight $\hat {\bm {w}}_{m,t}$ can be directly decided by the sign bit. When $\alpha_{t}\!=\!V$, no quantization takes place since we are dealing with full precision weights, i.e., $\hat {\bm {w}}_{m,t}\!=\!\bm {w}_{m,t}$. When $1 < \alpha_{t} < V$, the quantization delay incurred will increase based on the number of neurons in the neural network. For each neuron, the server will arithmetically perform the rounding, multiplication, and division operations according to (\ref{eq:BBNN1}).

\subsubsection{FL Model Transmission Delay}
To generate the global FL model that is aggregated by each quantized local FL model, each device must transmit $\hat{\bm w}_{m,t}$ to the server. To this end, we adopt an orthogonal frequency division multiple access (OFDMA) transmission scheme for quantized local FL model transmission. In particular, the server can allocate a set $\mathcal{U}$ of $U$ uplink orthogonal resource blocks (RBs) to the devices for quantized weight transmission. Let $W$ be the bandwidth of each RB and $P$ be the transmit power of each device. The uplink channel capacity between device $m$ and the server over each RB $i$ is $c_{m,t}\left(u_{m,t}\right)= u_{m,t} W{\log _2}\left(\!1\!+\! {\frac{{{P}{ h_{m,t}}}}{{ \sigma^2_{\emph{N}} }}} \!\right)\!$
where $u_{m,t}\in \{ 0,1\}$ is the user association index, $h_{m,t}$ is the channel gain between device $m$ and the server, and $\sigma^2_{\emph{N}}$ represents the variance of additive white Gaussian noise. Then, the uplink transmission delay between device $m$ and the server is $l_{m,t}^{\rm T}\left(u_{m,t},\alpha_{t}\right) =\frac{ D\alpha_{t}}{c_{m,t}\left(u_{m,t} \right)}$ where $D\alpha_{t}$ is the data size of the quantized FL parameters $\hat{\bm w}_{m,t}$.

Since the server has enough computational resources and sufficient transmit power, we do not consider the delay used for global FL model quantization and transmission. Thus, the time that the devices and the server require to jointly complete the update of their respective local and global FL models at iteration $t$ is given by
\begin{equation}\label{eq:sumdelay}
\begin{aligned}
&l_{t}\!\left(\bm u_{t},\!\alpha_{t}\right)\! =\! \mathop {\max }\limits_{m \in \mathcal{M}} u_{m,t}\!\left( l^{\rm C}_{m,t}(\alpha_{t})\!+l^{\rm Q}_{m,t}(\alpha_{t})\!+l_{m,t}^{\rm T}\!\left(u_{m,t},\!\alpha_{t}\right)\right).
\end{aligned}
\end{equation}
Here, $u_{m,t}=0$ implies that device $m$ will not send its quantized local FL model to the server, and thus not cause any delay.

\subsection{Problem Formulation}
The goal is to minimize the FL training loss while meeting a delay requirement on FL completion per iteration. This minimization problem involves jointly optimizing the device selection scheme and the quantization scheme, which is formulated as follows:
\begin{equation}\label{eq:max1}
\begin{split}
\!\!\!\!\!\!\!\!\!\!\!\!\!\!\!\!\!\!\!\!\!\!\!\!\!\!\!\!\!\!\!\!\!\!\!\!\!\!\!\!\!\!\!\!\!\!\!\!\!\!\!\!\!\!\!\!\!\!\!\!\!\!\!\!\!\!\!\!\!\!\!\!\! \mathop {\min}\limits_{{\bm U},\bm \alpha} \;\; F(\bm g({\bm u}_{T},\bm \alpha)),
\end{split}
\end{equation}
\vspace{-0.6cm}
\begin{align}\label{c1}
\setlength{\abovedisplayskip}{-20 pt}
\setlength{\belowdisplayskip}{-20 pt}
&\;\;\rm{s.t.}\;\;\scalebox{1}{${ u_{m,t}} \! \in \!  \left\{ {0,1} \right\},\alpha \in [0,V] \;{\rm and}\; \alpha \in \mathbb{N}^+,\forall {m} \in \mathcal{M},\forall {t} \in \mathcal{T},$}\tag{\theequation a}\\
&~~~~~~~\sum\limits_{m=1}^M u_{m,t}  \leqslant U,\;\forall m \in \mathcal{M},\forall {t} \in \mathcal{T},\tag{\theequation b}\\
&~~~~~~~ l_{t}\left(\bm u_{t},\bm \alpha\right)\leqslant \Gamma,\forall m \in \mathcal{M},\forall {t} \in \mathcal{T},\tag{\theequation c}
\end{align}
where ${\bm U}= [\bm u_{1},\ldots,\bm u_{t},\ldots,\bm u_{T}]$ is a device selection matrix over all iterations with $\bm u_{t}=[u_{1,t},\ldots,u_{M,t}]$ being a user association vector at iteration $t$, $\bm \alpha=[\alpha_1,\ldots,\alpha_{t},\ldots,\alpha_{T}]$ is a quantization precision vector of all devices for all iterations, and $\mathcal{T} = \{1,...,T\}$ is the training period. $\Gamma$ is the delay constraint for completing FL training per iteration, and $T$ is a large constant to ensure the convergence of FL. In other words, the number of iterations that FL needs to converge will be less than $T$. (\ref{eq:max1}a) indicates that each device can quantize its local FL model and can only occupy at most one RB for FL model transmission. (\ref{eq:max1}b) ensures that the server can only select at most $U$ devices for FL model transmission per iteration. (\ref{eq:max1}c) is a constraint on the FL training delay per iteration.

The problem in (\ref{eq:max1}) is challenging to solve by conventional optimization algorithms due to the following reasons. First, as the central controller, the server must select a subset of devices to collect their quantized local FL models for aggregating the global FL model. However, each local FL model that is generated by each device depends on the characteristics of the local dataset. Without such information related to the datasets, the server cannot determine the optimal device selection and quantization scheme for minimizing the FL training loss. Second, as the stochastic gradient decent method is adopted to generate each local FL model, the relationship between the training loss and device selection as well as quantization scheme cannot captured by the server via conventional optimization algorithms. This is because the stochastic gradient decent method enables each device to randomly select a subset of data samples in its local dataset for local FL model training, and hence, the server cannot directly optimize the training loss of each device. To tackle these challenges, we propose a model based RL algorithm that enables the server to capture the relationship between the FL training loss and the chosen device selection and quantization scheme. Based on this relationship, the server can proactively determine $\bm u_{t}$ and $\alpha_{t}$ so as to minimize the FL training loss.

\section{Optimization Methodology}

In this section, a model based RL approach for optimizing the device selection scheme ${\bm U}$ and the quantization scheme $\bm \alpha$ in (\ref{eq:max1}) is proposed. Compared to traditional model free RL approaches that continuously interacts with edge devices to learn the device selection and quantization schemes, model based RL approaches enable the server to mathematically model the FL training process thus finding the optimal device selection and quantization  scheme based on the learned state transition probability matrix. Next, we first introduce the components of the proposed model based RL method. Here, a linear regression method is used to learn the dynamic environment model in RL approach. Then, we explain the process of using the proposed model based RL method to find the global optimal ${\bm U}$ and $\bm \alpha$. Finally, the convergence and complexity of the proposed RL method is analyzed.
\subsection{Components of Model Based RL Method}

The proposed model based RL method consists of six components: a) agent, b) action, c) states, d) state transition probability,  e) reward, and f) policy, which are specified as follows:

\begin{itemize}
	\item \textbf{Agent}: The agent that performs the proposed model based RL algorithm is the server. In particular, at each iteration, the server must select a suitable subset of devices to transmit their local FL models and determine the number of bits used to represent each element in FL model matrix.
	
	\item \textbf{Action}: An action of the server is ${\bm a_t} = [{\bm u_t}, \alpha_t] \in \mathcal{A} $ that consists of the device selection scheme ${\bm u_t}$ and the quantization scheme $\alpha_t$ of all device at iteration $t$ with $\mathcal{A}$ being the discrete sets of available actions.
	
	\item \textbf{States}: The state is $s_t=F(\bm g_{t}) \in \mathcal{S} $ that measures the performance of global FL model at iteration $t$ with $F(\bm g_{t})$ being the FL training loss and $\mathcal{S}$ being the sets of available states.
	
	\item \textbf{State Transition Probability}: The state transition probability $P\left(s_{t+1}|s_t,\bm a_t \right)$ denotes the probability of transiting from state $s_t$ to state $s'_t$ when action $\bm a_t$ is taken, which is given by
	\begin{equation}\label{eq:reward1}
	 	P\left(s_{t+1}|s_t,\bm a_t \right)={\rm Pr}\{s_{t+1}=s_t'|s_{t},\bm a_t\}.
	\end{equation}
Here, we need to note that in model free RL algorithms, the server does not know the values of a state transition probability matrix. However, in our work, we analyze the convergence of FL and estimate the FL training parameters in the FL convergence analytical results so as to calculate the state transition probabilities. Using the state transition probability matrix can reduce the interactions between the server and edge devices thus improving the convergence speed of RL.

    \item \textbf{Reward}: Based on the current state $s_t$ and the selected action ${\bm a_t}$, the reward function of the server is
	given by
	\begin{equation}\label{eq:reward1}
		r_{} \left(\bm s_t,\bm a_{t}\right)=-F(\bm g({\bm u}_{t},\alpha_t)),
	\end{equation}
	where $F(\bm g({\bm u}_{t},\alpha_t))$ is the training loss at iteration $t$.
	Note that, $r_{} \left(\bm s_t,\bm a_{t}\right)$ increases as $F(\bm g({\bm u}_{t},\alpha_t))$ decreases, which implies that maximizing the reward of the server can minimize the FL training loss.
	
	\item \textbf{Policy}: The policy is the probability of the agent choosing each action at a given state. The model based RL algorithm uses a deep neural network parameterized by $\bm \theta$ to map the input state to the output action. Then, the policy can be expressed as $\bm \pi_{\bm \theta}\left(s_t,\bm a_t\right)=P(\bm a_t|s_t)$. 
	
\end{itemize}

\subsection{Calculation of State Transition Probability}

In this section, we introduce the process of calculating the state transition probability that is used to reduce the interactions between the server and edge devices thus improving the convergence speed of RL. To this end, we must analyze the relationship between $s_{t+1}$ and $(s_t,\bm a_t)$. First, we make the following assumptions, as done in \cite{APP}:

\begin{itemize}
	
	\item \textbf{Assumption 1}: The loss function $F(x)$ is $L-$smooth with the Lipschitz constant $L>0$, such that
	\begin{equation}\label{eq:Assumption1}
		\begin{aligned}
			||\nabla F(x)\!-\!\nabla F(y)||\!\leq\! L||x-y||.
		\end{aligned}
	\end{equation}
	
	\item \textbf{Assumption 2}: The loss function $F(x)$ is strongly convex with positive parameter $\mu$, such that
	\begin{equation}\label{eq:Assumption2}
		\begin{aligned}
			F(\bm g_{t+1})\geq F(\bm g_{t})+(\bm g_{t+1}\!-\!\bm g_{t})^T \nabla  F(\bm g_{t})\!+\!\frac{\mu}{2}||\bm g_{t+1}\!-\!\bm g_{t} ||.
		\end{aligned}
	\end{equation}
	
	\item \textbf{Assumption 3}: The loss function $F(x)$ is twice-continuously differentiable. Based on (\ref{eq:Assumption1}) and (\ref{eq:Assumption2}), we have
	\begin{equation}\label{eq:Assumption3}
		\begin{aligned}
			\mu {\bm I} \! \preceq\! \nabla^2 \!F({\bm g}_{t},\!\bm x_{mn},\bm y_{mn})\! \preceq\! L {\bm I}.
		\end{aligned}
	\end{equation}
	
	\item We also assume that 
	\begin{equation}\label{eq:Assumption4}
		\begin{aligned}
			\left\| \nabla f( {\bm g}_{t},\bm x_{mn},\bm y_{mn})\right\|^2 \!\leq\! \zeta_1 \!+\! \zeta_2\left\| \nabla F({\bm g}_{t})\right\|^2,
		\end{aligned}
	\end{equation}
	where $F({\bm g}_{t})\!=\!\frac{1}{N} \! \sum\limits_{m = 1}^M \!\sum\limits_{n = 1}^{N_m} \!f\!\left(\bm g_t,\bm x_{m,n},\bm y_{m,n}\right)$.
	
\end{itemize}

These assumptions can be satisfied by several widely used loss functions such as mean squared error, logistic regression, and cross entropy \cite{Assumption1}. These popular loss functions can be used to capture the performance of implementing practical FL algorithms for identification, prediction, and classification. Based on these assumptions, next, we first derive the upper bound of the improvement of the FL training loss at one FL training step under the non-i.i.d. setting. Then, we further analyze the relationship between the FL training loss improvement and the selected action (i.e., the relationship between $s_{t+1}$ and $s_t$ when $\bm a_t$ is given). Based on the analytical result, we can calculate the state transition probability $P\left(s_{t+1}|s_t,\bm a_t \right)$. To obtain the upper bound of the FL training loss improvement at one FL training step under the non-i.i.d. setting, we first define the degree of the non-i.i.d. data distribution.

\textbf{Definition 1}: 
The degree of non-i.i.d. in the global data distribution can be characterized by \cite{YZZ}:
\begin{equation}\label{eq:Noiid2}
\begin{aligned}
\epsilon = \sum\limits_{m = 1}^M \sum\limits_{n =1 }^{N_m} \frac{u_{m,t} \epsilon_{m} N_{m,t}}{\sum\limits_{m = 1}^M N_{m,t}} 
\end{aligned}
\end{equation}
where $\epsilon_{m} = \nabla  F(\bm g_{t}) - \nabla  \widetilde{F}_m(\bm g_{t})$ is the difference between the data distribution of device $m$ and the global data distribution. We also assume that $\left\| \nabla f( {\bm g}_{t},\bm x_{mn},\bm y_{mn}) + \epsilon_m \right\|^2 \!\leq\! \zeta_1 \!+\! \zeta_2\left\| \nabla F({\bm g}_{t})\right\|^2 + B \epsilon^2 $ for some positive $B$ with $F({\bm g}_{t})\!=\!\frac{1}{N} \! \sum\limits_{m = 1}^M \!\sum\limits_{n = 1}^{N_m} \!f\!\left(\bm g_t,\bm x_{m,n},\bm y_{m,n}\right)$.

Using Definition 1, we derive the upper bound of the FL training loss improvement at one FL training step under the non-i.i.d. setting.
\begin{lemma}
{\rm The FL training loss improvement over one iteration (i.e., the gap between $\mathbb{E}\left(F({\bm g}_{t+1})\right)$ and $\mathbb{E}\left(F({\bm g}_{t})\right)$) with a non-i.i.d. data distribution can be upper bounded as}
\begin{equation}\label{eq:L1}
\begin{split}
\mathbb{E}\left(F({\bm g}_{t+1})\right)&-\mathbb{E}\left(F({\bm g}_{t})\right)\\
&\leq  \mathbb{E}\left( \left(\hat {\bm g}_{t+1}- {\bm g}_{t}\right) \left(\nabla F({\bm g}_{t})-\epsilon \right) \right)+\frac{L}{2}\mathbb{E}\left(||\hat {\bm g}_{t+1}-{\bm g}_{t}||^2\right)+\frac{L}{2}\mathbb{E}\left(||\hat {\bm g}_{t+1}- {\bm g}_{t+1}||^2\right),
\end{split}
\end{equation}
{\rm where ${\bm g}_{t}$ and $\hat {\bm g}_{t}$ are short for ${\bm g}_{t}(\bm u_t,\alpha_t)$ and $\hat {\bm g}_{t}(\bm u_t,\alpha_t)$, respectively. $\mathbb{E}(\cdot)$ is the expectation with respect to the Rayleigh fading channel gain $h_{m,t}$ and quantization error.} 
\end{lemma}
\begin{IEEEproof}See Appendix A.
\end{IEEEproof}

From Lemma 1, we can see that, the upper bound of the FL training loss improvement at one iteration depends on $\hat {\bm g}_{t+1}(\bm u_{t+1},\alpha_{t+1})- {\bm g}_{t}(\bm u_{t},\alpha_{t})$ that is determined by the device selection vector $\bm u_t$ and quantization scheme $\alpha_t$. To investigate how an action $\bm a_t=[\bm u_t,\alpha_t]$ affects the state transition in the considered bitwidth FL algorithm with non-i.i.d. data distribution, we derive the following theorem:
\begin{theorem}
{\rm Given the user selection vector $\bm u_t$ and quantization scheme $\alpha_t$, the upper bound of $\mathbb{E}\left(F(\bm g_{t+1})\right)-\mathbb{E}\left(F(\bm g_{t})\right)$ in non-i.i.d. data distribution can be given by}
\begin{equation}\label{eq:L6}
\begin{split}
\mathbb{E}\left(F( {\bm g}_{t+1})\right)&-\mathbb{E}\left(F\left({\bm g}_{t}\right)\right)\\
&\leq\frac{1}{2L}\left(-1+\frac{4\left(N-A\right)^2\left(\mathbb{E}\left\|\Delta\left(\alpha_t\right)\right\|+1\right)\zeta_2}{N^2}\right)||\nabla F({\bm g}_{t})||^2\\
&\quad \!\! +\frac{\mathbb{E}\left\|\Delta\left(\alpha_t\right)\right\|+1}{2L}\left(\frac{4\left(N-A\right)^2\left(\zeta_1+B\epsilon^2\right)}{N^2} +L^{2} \mathbb{E}\left\|\Delta\left(\alpha_t\right)\right\| \right)+\mathbb{E}\left(\Delta\left(\alpha_t\right)^2\right),
\end{split}
\end{equation}
{\rm where $A\!=\!\sum\limits_{m=1}^M \!\!u_{m,t} N_{m,t} $ represents the sum of all selected devices' data samples that are used to train their local models, $\Delta(\alpha_t)\!=\!\hat {\bm g}_{t}(\alpha_t)\!-{\bm g}_{t}$ is the quantization error of the global FL model that depends on the quantization scheme $\bm \alpha$, $\mathbb{E}\left\|\Delta(\alpha_t)\right\|\!=\!M 2^{-\alpha_t}$ is the unbiased quantization function defined in (\ref{eq:BBNN}).}  
\end{theorem}
\begin{IEEEproof}See Appendix B.
\end{IEEEproof}

From Theorem 1, we can see that, the relationship between $\mathbb{E}\left(F(\bm g_{t+1})\right)$ and $\mathbb{E}\left(F(\bm g_{t})\right)$ (i.e., $s_{t+1}$ and $s_t$) depends on the selected action $\bm a_t$ as well as the constants $1/L$, $\zeta_1$, $\zeta_2$, and $B\epsilon^2$. However, we do not know the values of $1/L$, $\zeta_1$, $\zeta_2$, and $B\epsilon^2$ since they are predefined in assumptions (15)--(19). To find the tightest bound in (21), we must find the values of $1/L$, $\zeta_1$, $\zeta_2$, and $B\epsilon^2$ so as to build the relationship between $s_{t+1}$ and $s_t$ and calculate the state transition probability $P\left(s_{t+1}|s_t,\bm a_t \right)$. To this end, a linear regression method \cite{Distances} is used to determine the values of $L$, $\zeta_1$, $\zeta_2$, and $B\epsilon^2$ since the relationship between $\mathbb{E}\left(F(\bm g_{t+1})\right)-\mathbb{E}\left(F(\bm g_{t})\right)$ and these constants are linear. The regression loss function defined as 
\begin{equation}\label{eq:regression loss}
	\begin{aligned}
		&\mathcal{J}(L, \zeta_1, \zeta_2, B\epsilon^2)\\
		&=\frac{1}{I}\sum\limits_{i = 1}^I \left( \left(\mathbb{E}\left(F(\bm g_{t+1})^{(i)} \right)-\mathbb{E}\left(F(\bm g_{t})^{(i)}\right) \right)-K \left( L,\zeta_1,\zeta_2,B\epsilon^2 | F(\bm g_{t})^{(i)},\bm a_t^{(i)},F(\bm g_{t+1})^{(i)}\right)\right)^2,
	\end{aligned}
\end{equation}
where $I$ is the number of real interactions between the server and edge devices used to estimate $1/L$, $\zeta_1$, $\zeta_2$, and $B\epsilon^2$. $K\!\left(\!L,\zeta_1,\zeta_2,B\epsilon^2 | F(\bm g_{t})^{(i)},\!\bm a_t^{(i)}\!,\!F(\bm g_{t+1})^{(i)} \right)$ is the upper bound of the FL training loss at one FL training step obtained in (\ref{eq:TT1}). $\bm b^{(i)}\!=\!(F(\bm g_{t})^{(i)}\!,\!\bm a_t^{(i)\!}\!,\!F(\bm g_{t+1})^{(i)})$ is the set of recorded pairs consisted of FL training loss and the selected action observed by the server and devices. $\bm b^{(i)}$ will be used to estimate the values of $1/L$, $\zeta_1$, $\zeta_2$, and $B\epsilon^2$. Specifically, given $\mathcal{B}\!=\!\left\{\bm b^{(0)}\!,\!\ldots\!,\!\bm b^{(i)}\!,\!\ldots\!,\!\bm b^{(I)}\right\}$, $L$, $\zeta_1$, $\zeta_2$, and $B\epsilon^2$ are updated using a standard gradient descent method
\begin{equation}\label{b1}
\begin{aligned}
L=L-\iota_L\frac{\partial \mathcal{J}( L, \zeta_1, \zeta_2, B')}{\partial L},{\kern 9pt}\zeta_1=\zeta_1-\iota_{\zeta_1}\frac{\partial \mathcal{J}(L, \zeta_1, \zeta_2, B')}{\partial \zeta_1},\\
{\kern 9pt}\zeta_2\!=\!\zeta_2\!-\!\iota_{\zeta_2}\frac{\partial \mathcal{J}(L, \zeta_1, \zeta_2, B')}{\partial \zeta_2},{\kern 9pt}B'=B'-\iota_{B'}\frac{\partial \mathcal{J}(L, \zeta_1, \zeta_2, B')}{\partial B'},
\end{aligned}
\end{equation}
where $B'=B\epsilon^2$. $\iota_L$, $\iota_{\zeta_1}$, $\iota_{\zeta_2}$, and $\iota_{B'}$ are learning rates for parameters $L$, $\zeta_1$, $\zeta_2$, and $B'$.

Given the values of $L$, $\zeta_1$, $\zeta_2$, and $B\epsilon^2$, the gap between $\mathbb{E}\!\left(F( {\bm g}_{t\!+\!1})\right)$ and $\mathbb{E}\!\left(F({\bm g}_{t})\right)$ can be estimated according to our upper bound. Based on the definition of the state, the state transition probability $P(s_t+1|s_t, \bm a_t)$ is given by
\begin{equation}\label{eq:TTYC}
	\begin{aligned}
		&P\left(s_{t+1}|s_t,a_t\right)=\ 
		\left\{ \begin{array}{l}
			1, {\rm if} \; s_{t+1}=s_{t}
			+K\left(L,\zeta_1,\zeta_2,B\epsilon^2 | F(\bm g_{t})^{(i)},\bm a_t^{(i)},F(\bm g_{t+1})^{(i)} \right),\\
			0,  \;{\rm otherwise}.\\
		\end{array} \right.\
	\end{aligned}
\end{equation}
\subsection{Optimization of Device Selection and Quantization Scheme}

Having the state transition probability $P\left(s_{t+1}|s_t,\bm a_t \right)$, next, we introduce the optimization of $\bm \pi_{\bm \theta}$ so as to find the optimal device selection scheme $\bm u_t$ and quantization scheme $\alpha_t$. Optimizing $\bm \pi_{\bm \theta}$ for minimizing the FL training loss corresponds to minimizing 
\begin{equation}\label{modelbased}
\mathcal{L}(\bm \theta)=\!\sum\limits_{(s_t,\bm a_t) \in \bm \tau} P(s_0) \prod \limits_{t=1}^T \bm \pi_{\theta}(s_{t-1},\bm a_t)P(s_t|s_{t-1},\bm a_t) \sum\limits_{t = 1}^Tr_{} \left(\bm s_t,\bm a_{t}\right),
\end{equation}
where $\bm \tau=\{s_0,\bm a_0,\ldots,s_T,\bm a_T\}$ is the trajectory replay buffer.


Given (\ref{modelbased}), the optimization of policy network $\bm \theta$ is 
\begin{equation}
	\begin{aligned}\label{max_modelbased}
		\mathop {\max }\limits_{\bm \theta}\mathcal{L}(\bm \theta).
	\end{aligned}
\end{equation}

We update $\bm \pi_{\bm \theta}$ using a standard gradient descent method
\begin{equation}
	\begin{aligned}\label{update}
		\bm \theta=\bm \theta + \iota \nabla_{\bm \theta}\mathcal{L}(\bm \theta),
	\end{aligned}
\end{equation}
where $\alpha$ is the learning rate and the policy gradient is
\begin{equation}
	\begin{aligned}\label{update11}
\nabla_{\bm \theta}\mathcal{L}(\bm \theta)=
&\sum\limits_{(s_t,\bm a_t) \in \bm \tau} P(s_0) \prod \limits_{t=1}^T \bm \pi_{\theta}(s_{t-1},\bm a_t)P(s_t|s_{t-1},\bm a_t)\sum\limits_{t = 1}^T r \left(\bm s_t,\bm a_{t}\right)\\
=&\frac{1}{T} \sum\limits_{t=1}^T  r_{} \left(\bm s_t,\bm a_{t}\right) \nabla {\rm log} {\bm \pi}_{\bm \theta}(s_t,\bm a_t).
	\end{aligned}
\end{equation}

\subsection{Proposed Method for FL with Nonconvex Loss Functions}

In Sections III. A, B, and C, we proposed a novel model based RL to optimize the device selection and quantization scheme so as to minimize FL training loss. Here, we extend the designed RL for FL with non-convex loss functions. First, we derive the convergence of FL with non-convex loss functions. In particular, we first replace convex Assumptions 2 and 3 with the following conditions:

\textbf{Condition 1} \cite{ZAZ}: The gradient of the non-convex loss function $F(x)$ is bounded by a nonnegative constant $B$, i.e., $|| \nabla F(x)|| \leqslant C$.

\textbf{Condition 2} \cite{SSMC}: Function $F(x)$ is 
$\mu$-nonconvex such that all eigenvalues of $\nabla^2 F$ lie in $[-\mu,L]$, for some $\mu \in (0, L]$.

\textbf{Condition 3}\cite{SSMC}: The Hessian of the loss function $F(x)$ is $\gamma$-Lipschitz continuous, such that 
\begin{equation}\label{eq:AssumptionR2}
\begin{aligned}
	||\nabla^2 F(x)-\nabla^2 F(y)||\leqslant \gamma||x - y||.
\end{aligned}
\end{equation}

Together with Assumption 1, and Conditions 1 and 3, loss function $F(x)$ is $L_{\gamma}$-smooth for $L_{\gamma}= 4L + \gamma \iota C$ with $\iota \in [0,1/L]$ [Lemma 4.2, \cite{SSMC}]. Based on (\ref{eq:AssumptionR2}), Lemma 1 can be rewritten as
\begin{equation}\label{eq:LR1}
\begin{split}
\mathbb{E}\left(F\left( {\bm g}_{t+1}\right)\right)&-\mathbb{E}\left(F\left({\bm g}_{t}\right)\right) \\
&\leq   \mathbb{E}\left(\left(\hat {\bm g}_{t+1}- {\bm g}_{t}\right) \left(\nabla F({\bm g}_{t})-\epsilon \right) \right)+\frac{L_{\gamma}}{2}\mathbb{E}\left(\left\|\hat {\bm g}_{t+1}- {\bm g}_{t}\right\|^2\right)+\frac{L_{\gamma}}{2}\mathbb{E}\left(||\hat {\bm g}_{t+1}- {\bm g}_{t+1}||^2\right).
\end{split}
\end{equation}
Then, the convergence of our FL methodology with nonconvex loss functions is shown in the following theorem.

\begin{theorem}
{\rm Given the user selection vector $\bm u_t$ and quantization scheme $\alpha_t$, an upper bound  $\mathbb{E}\left(F(\bm g_{t+1})\right)-\mathbb{E}\left(F(\bm g_{t})\right)$ can be obtained as}
\begin{equation}\label{eq:TT1}
	\begin{aligned}
		&\mathbb{E}\left(F( {\bm g}_{t+1})\right)-\mathbb{E}\left(F({\bm g}_{t})\right)
		\leq\frac{1}{2L_{\gamma}}\left(-1+\frac{4(N-A)^2 \mathbb{E}\left(\left\|\Delta( \alpha_t)\right\|+1\right)\zeta_2}{N^2}\right)||\nabla F({\bm g}_{t})||^2 +\frac{M^2\Upsilon^4}{2L_{\gamma}}\\
		&\qquad \qquad \qquad \qquad  +\frac{\mathbb{E}\left\|\Delta(\alpha_t)\right\|+1}{2L_{\gamma}}\left(\frac{4(N-A)^2(\zeta_1+B\epsilon^2)}{N^2} +{L_{\gamma}}^2 \mathbb{E}\left\|\Delta(\alpha_t)\right\| \right)+\mathbb{E}\left(\Delta(\alpha_t)^2\right),\\
	\end{aligned}
\end{equation}
{\rm where $A\!=\!\sum\limits_{m=1}^M \!\!u_{m,t} N_{m,t} $ represents the sum of all selected devices' data samples, $\Delta(\alpha_t)\!=\!\hat {\bm g}_{t}(\alpha_t)\!-{\bm g}_{t}$ is the quantization error of the global FL model, and $\mathbb{E}\left\|\Delta(\alpha_t)\right\|\!=\!M 2^{-\alpha_t}$ is the unbiased quantization function.}
\end{theorem}
\begin{IEEEproof}See Appendix C.
\end{IEEEproof}

Given Theorem 2, the server is able to obtain the state transition probability $P(s_{t + 1}|s_t, a_t)$ via estimating ${L_{\gamma}}$, $\Upsilon$, $\zeta_1$, $\zeta_2$, $B\epsilon^2$. Given $P(s_{t + 1}|s_t, a_t)$, we can use the proposed model based RL to find the optimal device selection and quantization scheme.

\subsection{Implementation and Complexity}

Next, we first analyze the training process of the model based RL algorithm. 
To train the proposed model based RL, the server needs to collect the time consumption coefficient $\rho$, the frequency of the CPU $f$, the number of bits that can be processed by the CPU in one clock cycle $\vartheta$, the channel gain $h_{m,t}$, and the transmit power of each device $P$. These parameters are constant and can be obtained from devices. Meanwhile, the server already knows FL model meta-parameters such as the number of multiplication operations $N^{\rm C}$ and the number of neurons $D$, when it initializes the FL model. Additionally, during the initial $I$ FL training iterations, the server must first randomly select a subset of devices to participate in FL, obtaining their training loss values and the selected actions so as to estimate the values of $1/L$, $\zeta_1$, $\zeta_2$, and $B\epsilon^2$, and calculate $P\left(s_{t+1}|s_t,\bm a_t \right)$, as shown in (\ref{eq:regression loss})-(28). Then, using $P\left(s_{t+1}|s_t,\bm a_t \right)$, a model based RL algorithm is used to find the optimal $\alpha_t$ and $\bm u_{t}$ without any interactions between the server and devices. The entire process of training the proposed model based RL algorithm is shown in Algorithm 1.

\begin{algorithm}[t]\small
	\caption{Model-based RL for device selection and quantization optimization}
	\label{table}
	\begin{algorithmic} [1] 
		\REQUIRE The environment state $\mathcal{S}$, the action space $\mathcal{A}$.\\
		\ENSURE The device selection and quantization scheme.\\ 
		\STATE Initialize policy $\bm \pi_{\bm \theta}$, transition replay buffer $\mathcal{B}$, trajectory replay buffer $\bm \tau$.
		\FOR {iteration $i=1:I$}
        \STATE Randomly selects a subset of devices to generate the global FL model that are quantized into $\alpha_t$ bits.
        \STATE  Records $F(\bm g_{t})$, $F(\bm g_{t+1})$, $\alpha_t$, and device selection scheme $\bm u_t$ in $\mathcal{B}$.
		\ENDFOR 
		\STATE Estimate $1/L$, $\zeta_1$, $\zeta_2$, and $B\epsilon^2$ to construct $P\left(s_{t+1}|s_t,\bm a_t \right)$ using (23) based on the real transition in $\mathcal{B}$.  
		\FOR {iteration $i=1:H$}
		\STATE Sample initial state from $\mathcal{S}$, then use policy $\bm \pi_{\bm \theta}$ and learned $P\left(s_{t+1}|s_t,\bm a_t \right)$ to perform $T$ trajectories and update $\bm \tau$.
		\STATE Sample from $\bm \tau$, and update the current policy evaluation by solving Equation (28).
		\ENDFOR 
	\end{algorithmic}
\end{algorithm}

The computational complexity of the proposed algorithm lies in the calculation of the state transition probability $P\left(s_{t+1}|s_t,\bm a_t \right)$ by a standard gradient descent method as well as optimizing $\alpha_t$ and $u_{m,t}$ by the proposed model based RL algorithm, which is detailed as follows:

a) In terms of computational complexity of calculating $P\left(s_{t+1}|s_t,\bm a_t \right)$, the server needs to find the values of $1/L$, $\zeta_1$, $\zeta_2$, and $B\epsilon^2$ using a linear regression method. The regression loss function in (\ref{eq:regression loss}) is strongly convex and smooth. Hence, the fixed step-size gradient descent method at least updates $\mathcal{O}\!\left(\! \frac{||L^0-L^* \!||^2_2}{\varepsilon \iota_L} \!+\! \frac{||{\zeta_1}^0\!\!-{\zeta_1}^* \!||^2_2}{\varepsilon \iota_{\zeta_1}} \!+\! \frac{||{\zeta_2}^0\!\!-{\zeta_2}^* \!||^2_2}{\varepsilon \iota_{\zeta_2}} \!+\! \frac{||{B'}^0\!\!-{B'}^* \!||^2_2}{\varepsilon \iota_{B'}} \right)$ iterations for reaching the optimal $L^*$, ${\zeta_1}^*$, ${\zeta_2}^*$, $B'^*$ from initialized $L^0$, ${\zeta_1}^0$, ${\zeta_2}^0$, $B'^0$ with $\varepsilon$ error \cite{TFAS}.

b) The computational complexity of the proposed RL algorithm depends on the number of the parameters in the policy network $\bm \theta$ which depends on the size of action space $\mathcal{A}$ and the size of state space $\mathcal{S}$. In particular, the possible combinations of each $\bm a_t$ in action space $\mathcal{A}$ is to choose $i$ devices ($i\leq U$) from all $M$ devices to participate in quantized FL. Thus, the size of the possible device selections is $\sum\limits_{i = 1}^U C_M^i=\sum\limits_{i = 1}^U\frac{M!}{i!(M-i)!}$ and the number of quantization actions is $Y$. The state space $\mathcal{S}$ consists of the continuous values of loss function $F(\bm g_{t})$. To ensure a finite state space, the continuous loss function values is divided into $|\mathcal{S}|$ levels. Then, the size of state space $\mathcal{S}$ is $|\mathcal{S}|$. Therefore, the computational complexity of the proposed RL algorithm 
is $\mathcal{O}\left\{ Y |\mathcal{S}| \sum\limits_{i = 1}^U \frac{M!}{i!(M-i)!} \prod \limits_{k=2}^{K-1} H_k\right\},$
where $H_k$ is the number of the neurons in layer $k$ of the policy network $\bm \theta$.


\section{Numerical Evaluation}
 
For our simulations, we consider a circular network area having a radius $r = 1500 $ m with one server at its center serving $M = 15$ uniformly distributed devices. The other parameters used in simulations are listed in Table I, unless otherwise stated. For comparison purposes, we use three baselines: 
\begin{itemize}
	\item a) The binary FL scheme from \cite{SIGNFL} that enables the server to randomly select a subset of devices to cooperatively train the FL model at each iteration. Each parameter in the trained FL model is quantized into one bit.
    \item b) An FL algorithm that enables the server to randomly select a subset of devices to cooperatively train the FL model in full-precision (i.e., without quantization), which can be seen as a standard FL \cite{JLD95}.
    \item c) An FL algorithm that optimizes the device selection and quantization schemes using a model free RL method \cite{PCCZ7}.  For c), a policy gradient-based RL update is employed to learn the state transition probabilities.
\end{itemize}


\begin{table}
	\centering
	\vspace{-0.3cm}
	\renewcommand\arraystretch{1}
	\caption{Simulation Parameters} 
	\vspace{-0.2cm}
	\small
	\setlength{\tabcolsep}{0.9mm}{
		\begin{tabular}{|c|c|c|c|c|c|}
			\hline
			\textbf{Parameters}&\textbf{Values}&\textbf{Parameters}&\textbf{Values}&\textbf{Parameters}&\textbf{Values}\\
			\hline
			\emph{M}& 15 &\emph{U}& 6 & $N_m$& 200\\
			\hline
			\emph{W}& 15 kHz &$P$& 0.5 W &$\sigma^2_N$& -174 dBm \\
			\hline
			$K$& 8 & $I$ & 20 &$\Gamma$&1 s\\
			\hline
			$T$& 1000 & $f$ & 3.3 GHz & $B $& 64 \\
			\hline
			$ D $ & 217728 & $\rho$ & $2.8 \times 10^{6}$ & $\iota$& 0.02\\
			\hline
			$\iota_{L}$& 0.02 & $\iota_{\zeta_1}$& 0.02 &$\iota_{\zeta_2}$& 0.02 \\
			\hline
	\end{tabular}}
	\vspace{-0.4cm}
\end{table}

\subsection{Datasets and ML Models}

\begin{figure}[t]
\centering
\includegraphics[width=8.9cm]{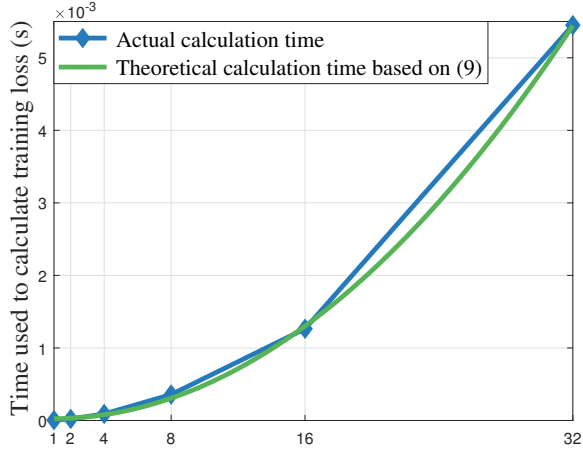}
\vspace{-1.2cm}
\caption{Calculation time of low bitwidth federated learning vs. the quantization precision.}
\label{inferance_time}
\end{figure}

We consider two popular ML tasks: handwritten digit identification on the MNIST dataset \cite{MNIST}, and image classification on the CIFAR-10 dataset \cite{CIA10}. The quantized FL algorithm that is used for handwritten digit identification consists of three full-connection layers. The total number of model parameters in the used fully-connected neural network (FNN) is 217728 ($= 28 \times 28 \times 256 + 256 \times 64 + 64 \times 10$). To verify the feasibility of the proposed calculation time model in (9), we first simulate the actual calculation time using the clock module in GEneral Matrix Multiply (GEMM) \cite{YM}, as shown in Fig. \ref{inferance_time}. Fig. \ref{inferance_time} shows that the actual calculation time is almost the same as the theoretical calculation time in (9).

The quantized FL algorithm that is used for image classification consists of three convolutional layers and two full-connection layers. In the used convolutional neural network (CNN), the size of the convolutional kernel is $5 \times 5$ and the total number of model parameters in CNN is 116704 ($= 5\times 5\times(3 \times 32 + 32 \times 32 + 32 \times 64) + 576 \times 64 \times64 + 10$). 

For both datasets, we will consider two cases of data distributions across clients: (i) non-i.i.d., where each client is allocated samples from only $3$ of $10$ labels; and (ii) i.i.d., where each client is allocated samples from all labels. All FL algorithms are considered to be converged when the value of the FL loss variance calculated over 20 consecutive iterations is less than 0.001.

\subsection{Convergence Performance Analysis}

\begin{figure}[t]
\centering
\begin{minipage}[t]{0.49\textwidth}
\centering
\includegraphics[width=8.9cm]{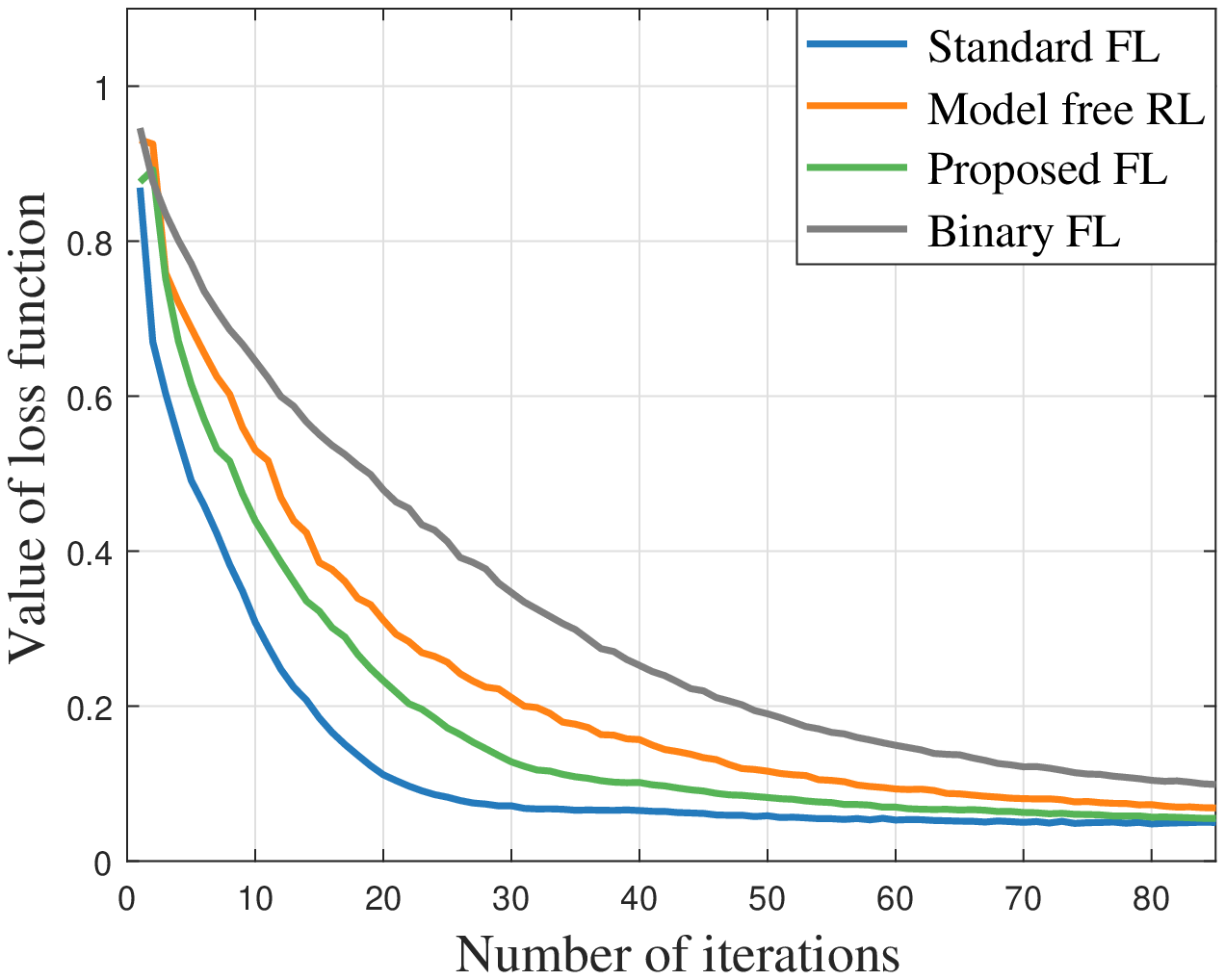}
\vspace{-1.2cm}
\caption{Training loss vs. the number of iterations.}
\label{converge_device1}
\end{minipage}
\begin{minipage}[t]{0.49\textwidth}
\centering
\includegraphics[width=8.9cm]{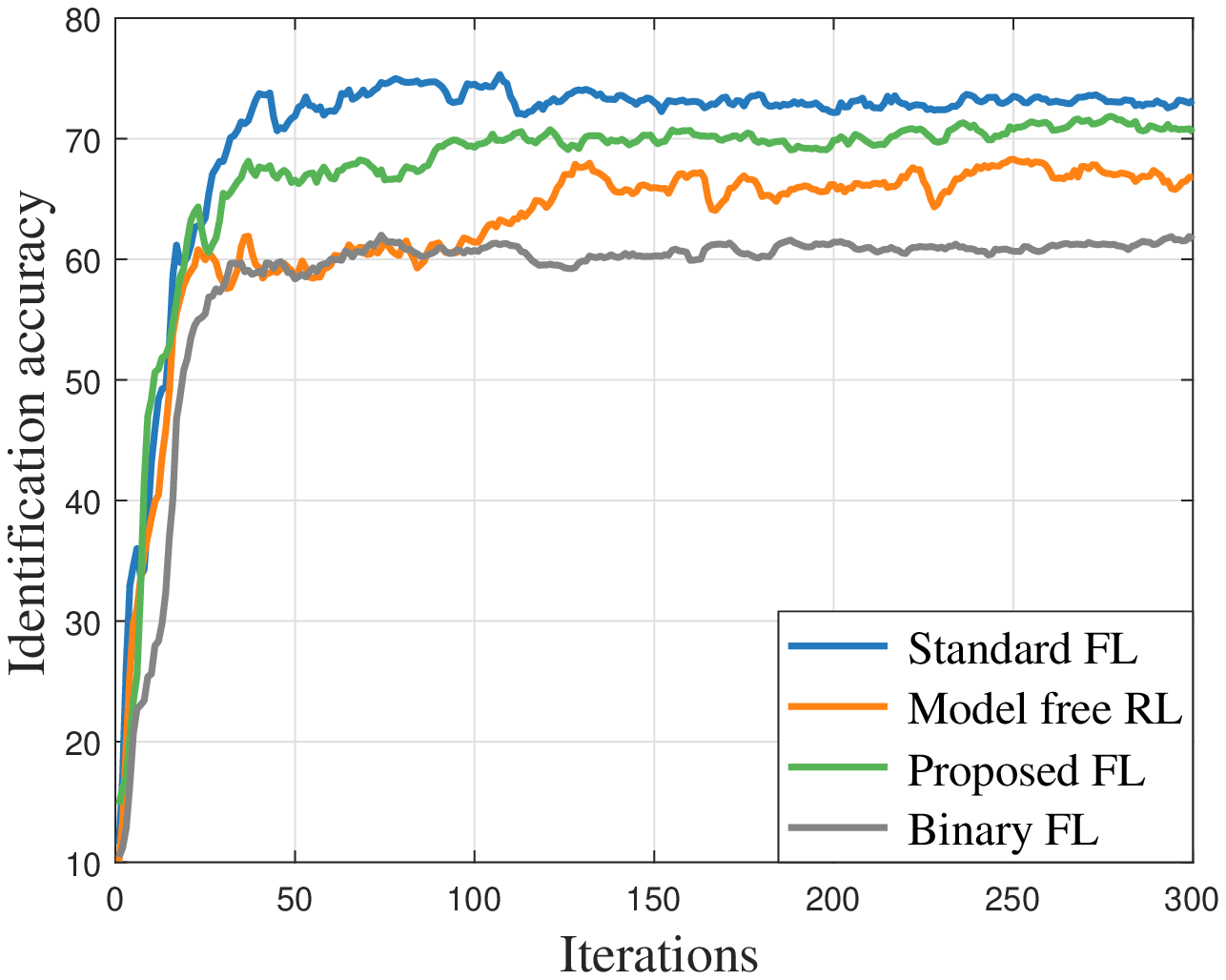}
\vspace{-1.2cm}
\caption{Identification accuracy vs. number of iterations.}
\label{noiid_acc}
\end{minipage}
\vspace{-0.8cm}
\end{figure}

Fig. \ref{converge_device1} shows how the FL training loss changes as the number of iterations varies for MNIST. From Fig. \ref{converge_device1}, we can see that, the proposed model based RL algorithm can reduce the number of iterations needed to converge by $14\%$ and $24\%$ compared to the model free RL method and the binary FL method, respectively. This is due to the fact that the proposed method enables the server to estimate the FL training parameters in the first few iterations so as to model the FL training process mathematically thus reducing the number of iterations required to converge.

Fig. \ref{noiid_acc} gives the accuracy plot corresponding to Fig. \ref{converge_device1}. From this figure, we can see once again that, the proposed algorithm obtains a noticeable improvement in convergence speed compared with model free RL in the non-i.i.d. case. This implies that our proposed algorithm models the FL training process effectively in the non-i.i.d. case via estimating the key meta-parameters that lead to speeding up the convergence. Fig. \ref{noiid_acc} also shows our algorithm comes within 2\% of the accuracy obtained by standard FL at convergence due to quantization errors. In contrast, by optimizing the device selection and quantization precision, our methodology reduces the necessary bitwidth by 68\% and the number of devices participating in each round by 30\%.

\begin{figure}[t]
\centering
\begin{minipage}[t]{0.49\textwidth}
\centering
\includegraphics[width=8.9cm]{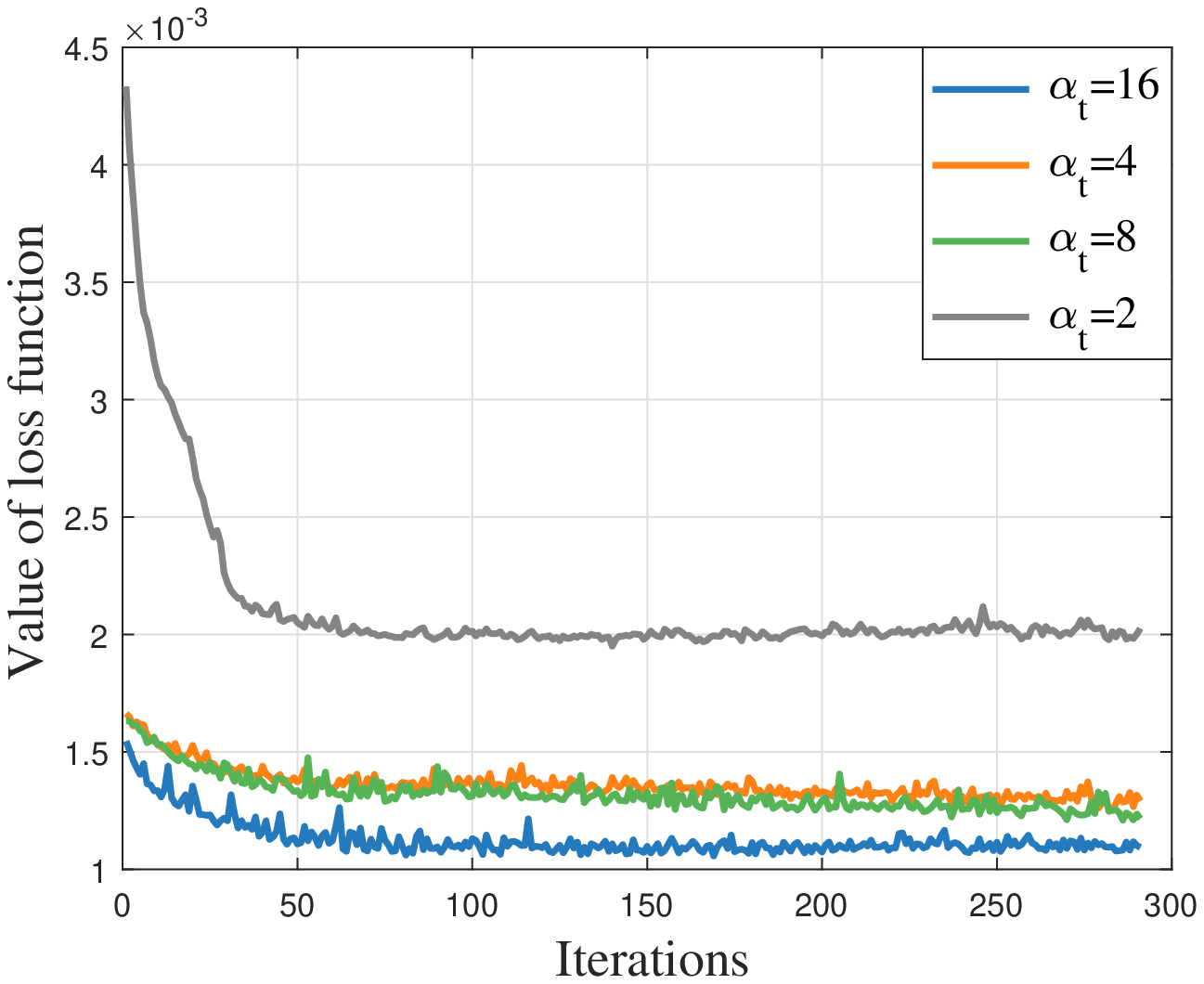}
\vspace{-1.2cm}
\caption{Training loss vs. number of iterations.}
\label{CNN_LOSSR}
\end{minipage}
\begin{minipage}[t]{0.49\textwidth}
\centering
\includegraphics[width=9cm]{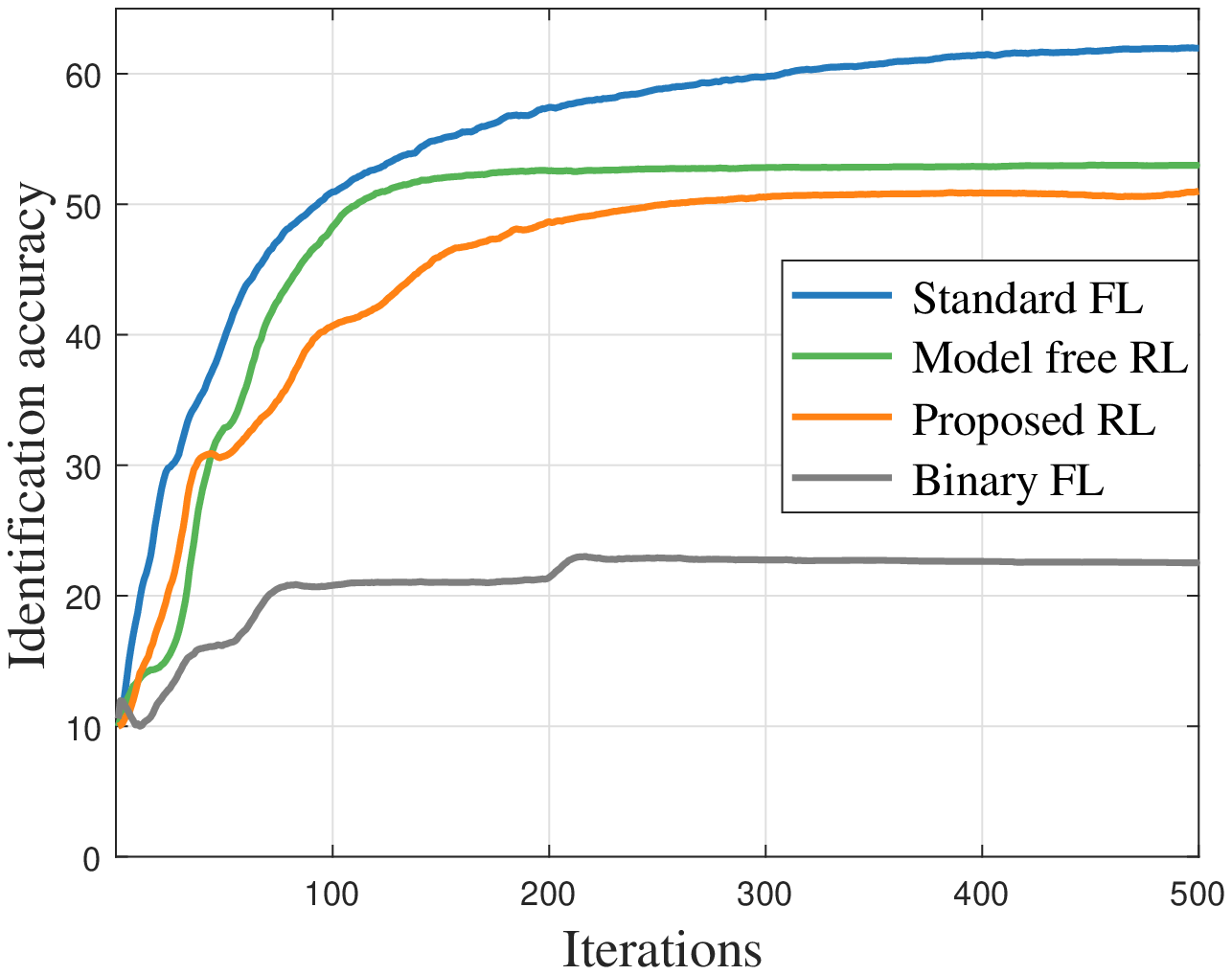}
\vspace{-1.2cm}
\caption{Identification accuracy vs. number of iterations.}
\label{CNNcase1}
\end{minipage}
\vspace{-0.8cm}
\end{figure}

Fig. \ref{CNN_LOSSR} shows how the FL training loss changes as the number of iterations varies on the CIFAR-10 dataset. We can see that the value of the loss function decreases as the number of iterations increases, and as the quantization bitwidth increases, the value of the loss function decreases. This is because as the quantization bitwidth increases, the introduced error resulting from the quantization decreases, which enables the trained FL model achieves a better performance in terms of training loss.

Fig. \ref{CNNcase1} shows how the identification accuracy of all considered algorithms changes as the number of iterations varies on the CIFAR-10 dataset in the non-i.i.d. case. We see that our proposed methodology can achieve up to 22\% improvement in terms of the number of iterations required to converge compared to the model free RL method. Similar to the previous figures, this demonstrates the advantage of the server estimating the associated FL model parameters based on information captured during the training process, thus optimizing the FL training process through minimal interaction with each device. Fig. \ref{CNNcase1} also shows that the binary FL algorithm (i.e., when the weights of CNN are binary) can only achieve 21\% identification accuracy. This is due to the fact that the binary FL neither has a pre-training process \cite{AKD} nor uses full-precision scale factors \cite{ICVV} to recover the full-precision model, again emphasizing the benefit of optimizing the quantization precision.

\subsection{Training Accuracy and Latency Comparison}

\begin{figure}[t]
	\centering
	\setlength{\belowcaptionskip}{-0.45cm}
	\vspace{-0.1cm}
	\includegraphics[width=16cm]{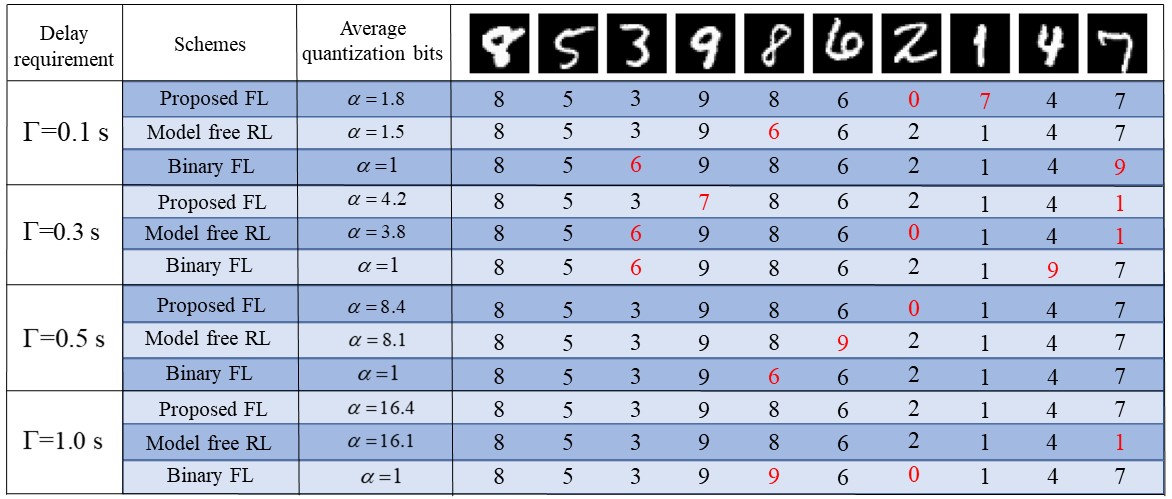}
	\vspace{-0.4cm}
	\centering
	\caption{An example of implementing quantized FL for handwritten digit identification.}
	\vspace{-0.5cm}
	\label{example1}
\end{figure}

Fig. \ref{example1} shows one example of implementing the proposed FL algorithm for 40 handwritten digit identification. From this figure, we see that, as the delay requirement $\Gamma$ for completing each FL training iteration increases, the average quantization bits $\alpha$ and the identification accuracy increase. This is because as $\Gamma$ increases, the time that can be used for training and transmitting local FL parameters in the selected devices increases thus resulting in an improvement of $\alpha$ and identification accuracy. From Fig. \ref{example1}, we can see that, for 40 handwritten digit identification, the proposed algorithm correctly identifies 35 handwritten digits. In contrast, the model free RL identifies 34 handwritten digits and the binary FL correctly identifies 33 handwritten digits. This is because the proposed FL algorithm can mathematically model the FL training process by obtaining the transition probability so as to find out the optimal device selection and quantization scheme for achieving a higher identification accuracy.

\begin{figure}[t]
\centering
\begin{minipage}[t]{0.49\textwidth}
\centering
\includegraphics[width=8.9cm]{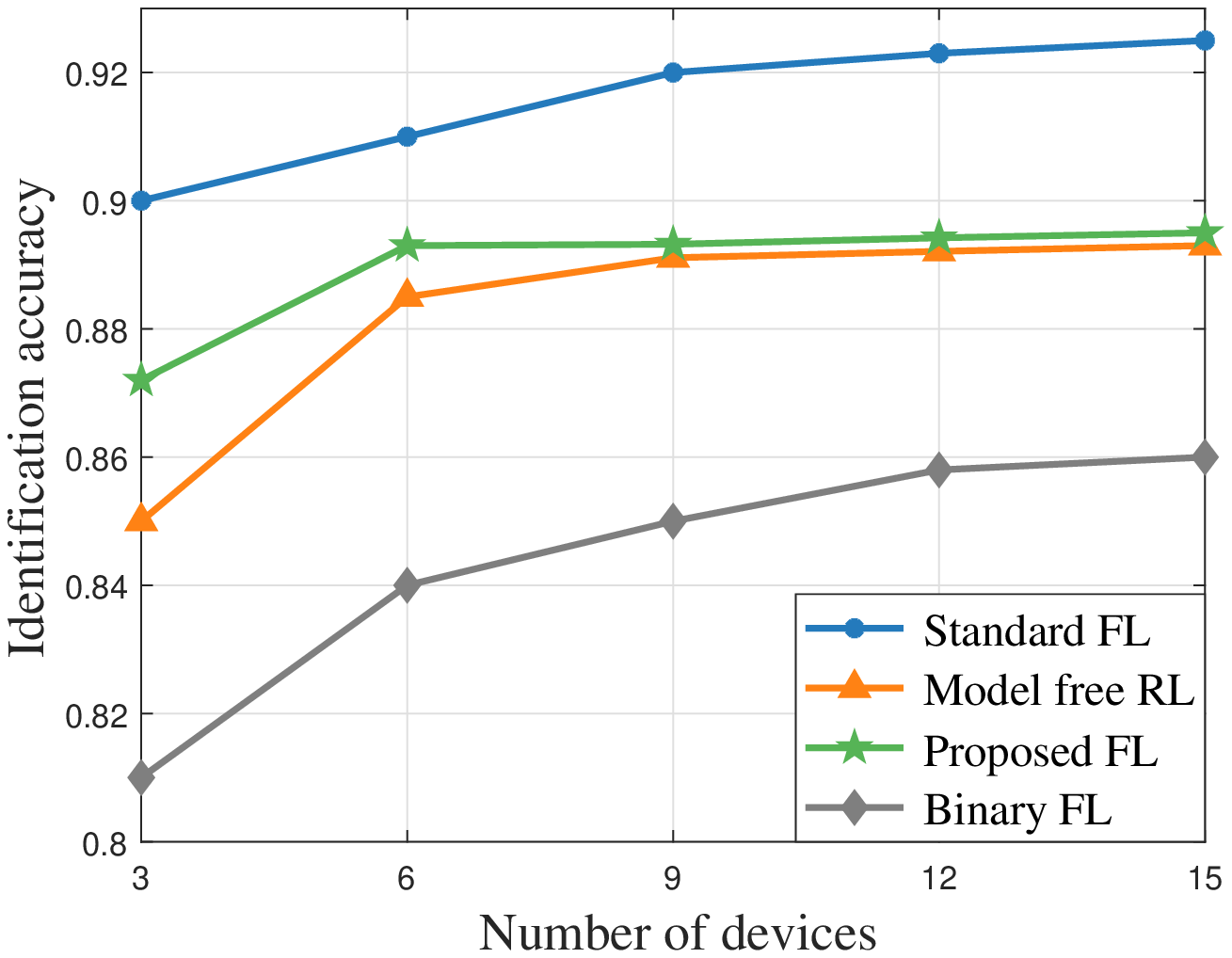}
\vspace{-1.2cm}
\caption{Identification accuracy vs. number of devices.}
\label{interval2}
\end{minipage}
\begin{minipage}[t]{0.49\textwidth}
\centering
\includegraphics[width=8.9cm]{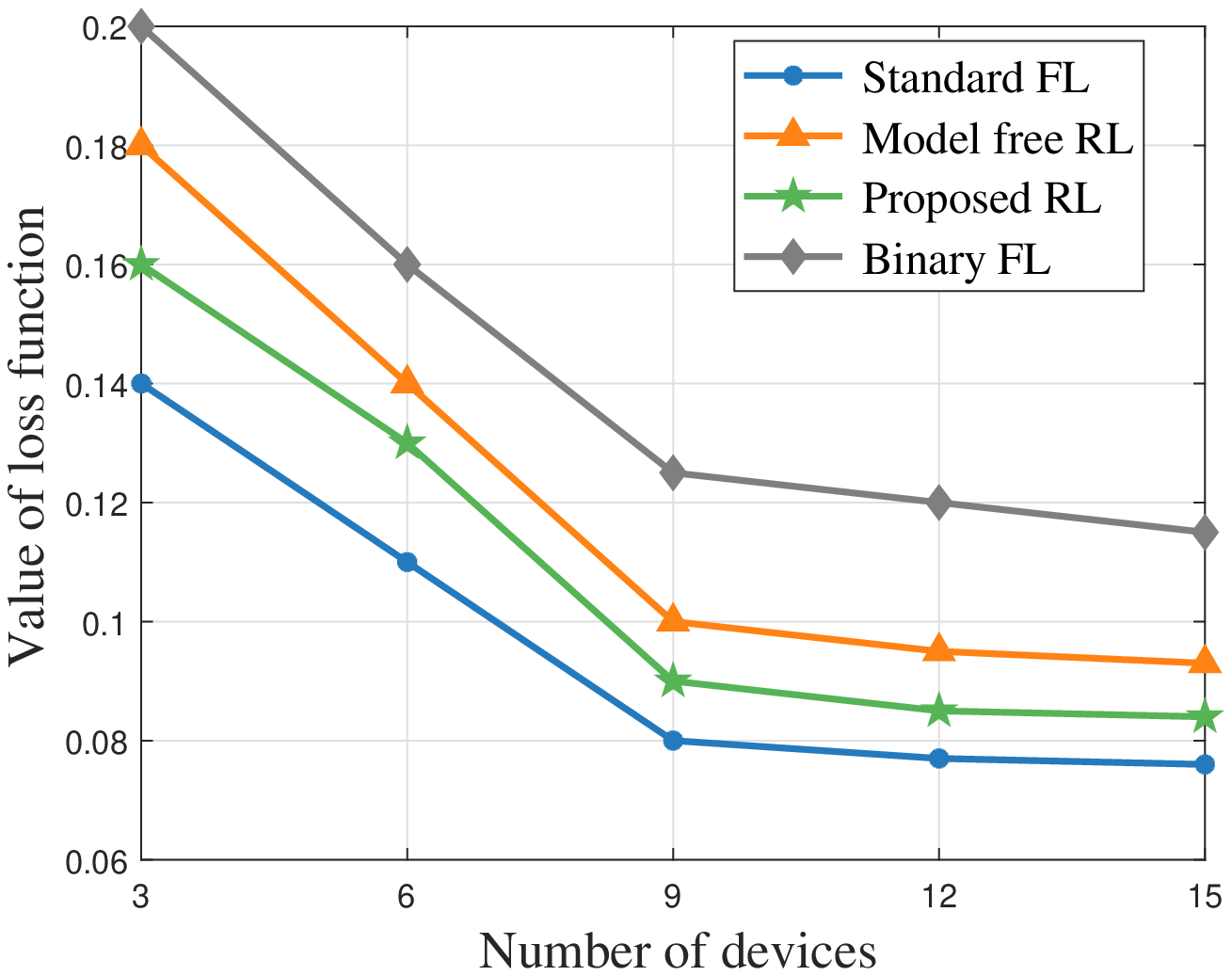}
\vspace{-1.2cm}
\caption{Training loss vs. number of devices.}
\label{interval1}
\end{minipage}
\vspace{-0.5cm}
\end{figure}

Fig. \ref{interval2} shows how the identification accuracy changes as the number of devices varies on the MNIST dataset, in the non-i.i.d. case. From Fig. \ref{interval2}, we see that the proposed model based RL method can improve the identification accuracy by up to 6\% compared with binary FL when the number of devices is small, while it is closer to 3\% when the number of devices grows larger. Our method is able to obtain a stronger robustness against the number of devices than model-free RL and binary FL through analyzing the relationship among the FL training loss at different iterations, thus finding a better FL training policy especially when less information is available. Fig. \ref{interval2} also shows that the performance of model-free RL almost converges to our method as the number of devices increases (in this case, once it reaches 9), This is because a smaller number of devices results in a smaller number of data samples available throughout the system for training the RL policy and FL model. One of the advantages of our model-based methodology is that it enables the server to model the FL training process using limited device-to-server interactions, which translates to lower data requirements. As the number of devices continues to increase, the considered network will have sufficient data samples for policy learning even with a model-free RL approach. Thus, the identification accuracy of the proposed method and that of the model-free RL become the same for larger $M$.

Fig. \ref{interval1} gives the training loss corresponding to the testing accuracy plot in Fig. \ref{interval2}. The result is consistent with what we observe in Fig. \ref{interval2}: as the number of devices increases, all considered algorithms have more data available for training, which gives a lower training loss. On the other hand, when there are fewer devices ($M < 9$), the training loss increases more quickly.

\begin{figure}[t]
	\centering
	\setlength{\belowcaptionskip}{-0.45cm}
	\vspace{-0.1cm}
	\includegraphics[width=16.5cm]{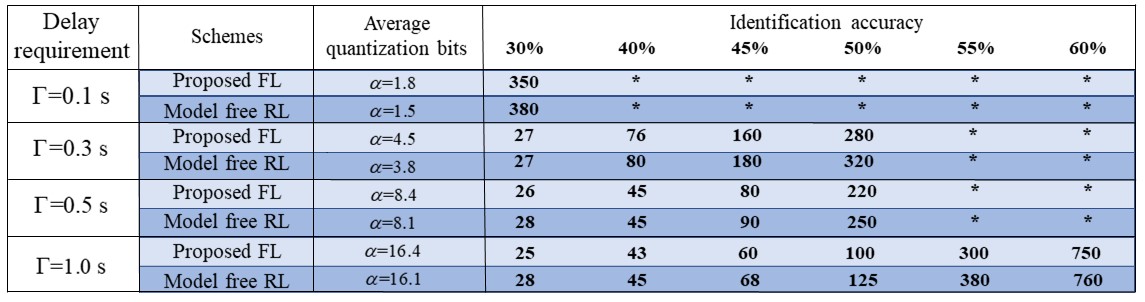}
	\vspace{-1cm}
	\centering
	\caption{An example of implementing quantized FL for CIFAR-10 identification.}
	\vspace{-0.7cm}
	\label{CNN_delay}
\end{figure}

Fig. \ref{CNN_delay} shows how the identification accuracy of the proposed FL framework changes as the delay requirement varies. This figure is simulated using CIFAR-10 dataset. From Fig. \ref{CNN_delay}, we see that, as the delay requirement increases, the identification accuracy of all considered learning algorithms increases. This is because that as the delay requirement increases, all considered learning algorithms enables the selected devices to fully utilize the training and transmitting time to perform FL framework, which results in an increase of average quantization bits and achievable accuracy. Fig. \ref{CNN_delay} also shows that as the average quantization bits $\alpha$ decreases, the number of iterations required to reach a fixed achievable accuracy increases slightly. This is due to the fact that as $\alpha$ decreases, the quantization error increases, which decreases the accuracy for modeling FL tranining process. However, with a decrease of $\alpha$, the time used to perform FL training at each iteration decreases and thus, the total time used to reach a fixed achievable accuracy decreases rapidly, which implies that the time used for training the proposed quantized FL framework decreases.

\begin{figure}[t]
\centering
\begin{minipage}[t]{0.50\textwidth}
\centering
\includegraphics[width=9cm]{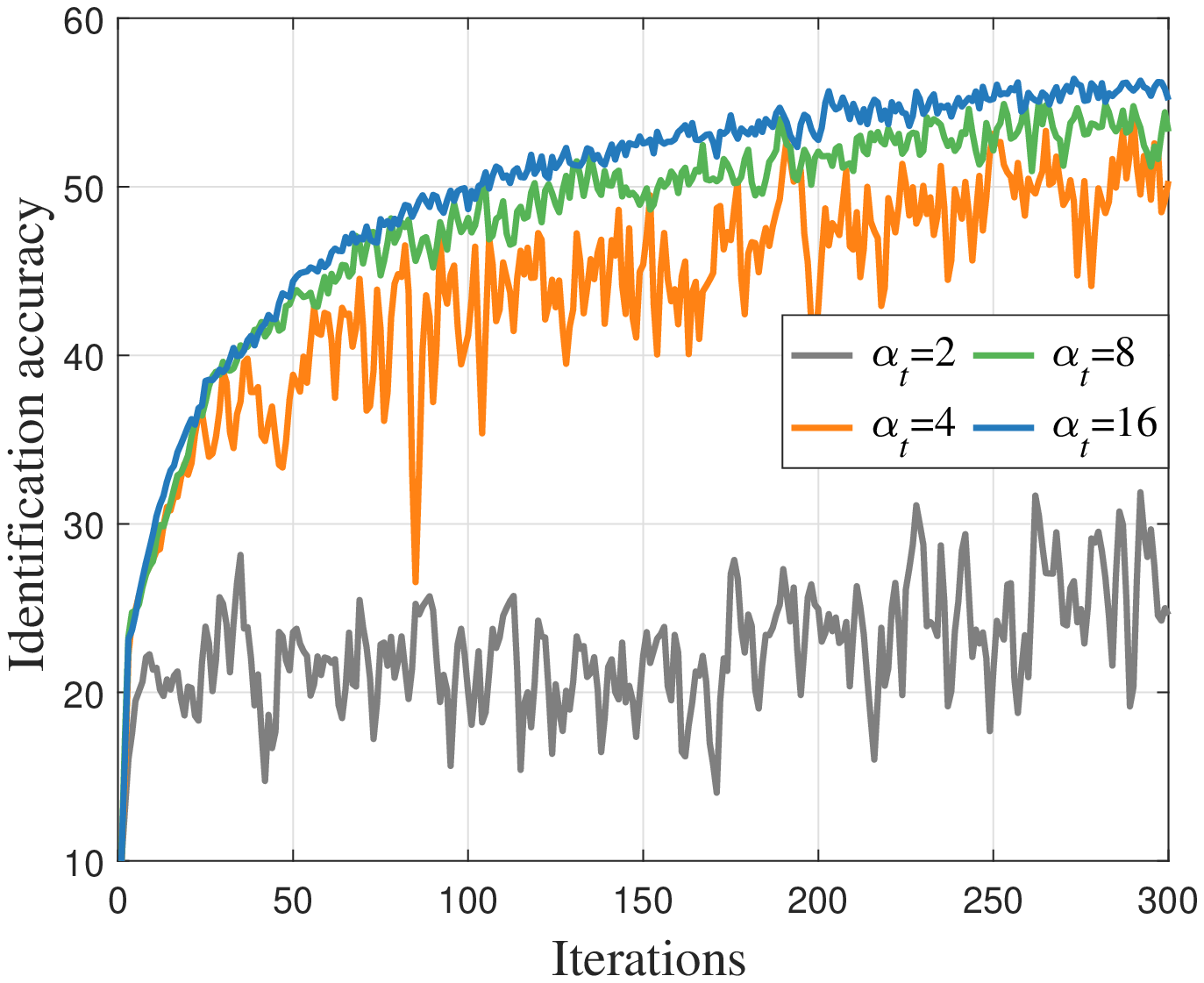}
\vspace{-1.2cm}
\caption{Identification accuracy vs. number of iterations.}
\label{changeBit}
\end{minipage}
\begin{minipage}[t]{0.49\textwidth}
\centering
\includegraphics[width=9.2cm]{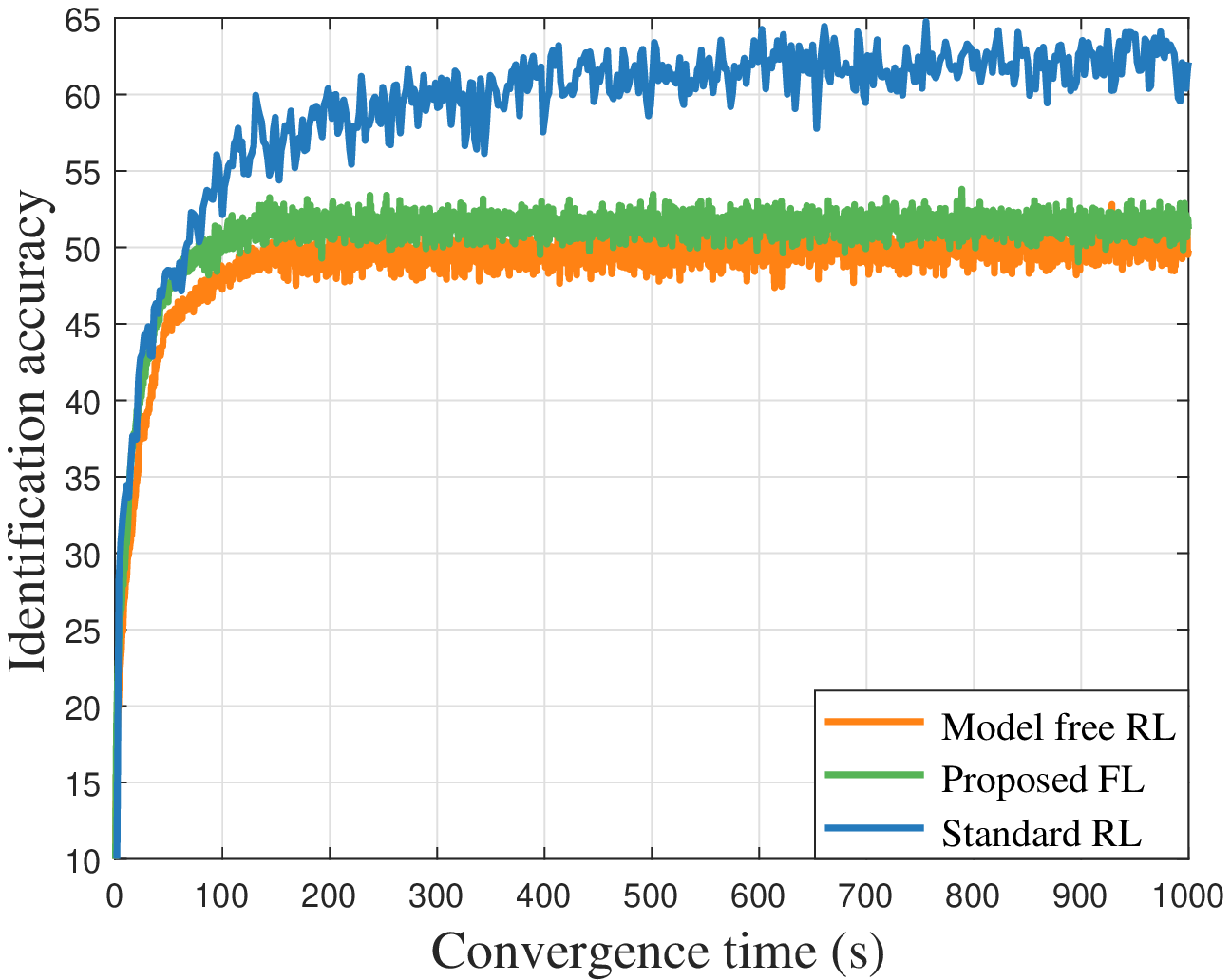}
\vspace{-1.2cm}
\caption{Identification accuracy vs. convergence time.}
\label{runtime}
\end{minipage}
\vspace{-0.5cm}
\end{figure}

In Fig. \ref{changeBit}, we show how the identification accuracy of the proposed FL algorithm changes as the number of iterations varies. This figure is simulated using CIFAR-10 dataset. From Fig. \ref{changeBit}, we see that, as the number of iterations increases, the identification accuracy of the proposed algorithms with different $\alpha$ first increases and, then remains unchanged. This is because FL algorithms converge. From Fig. \ref{changeBit}, we can also see that as $\alpha$ increases, the identification accuracy of the proposed algorithm increases. This is due to the fact that as $\alpha$ increases, the quantization error decreases. Fig. \ref{changeBit} also shows that as $\alpha$ decreases, the instability of the proposed algorithm increases. This is because the quantization error in the weights of the convolution kernel can significantly affect the result of convolution operation, thus a decrease of $\alpha$ will result in a degeneration of identification accuracy in CNN.

Fig. \ref{runtime} shows how the identification accuracy changes as the convergence time varies. This figure is simulated using CIFAR-10 dataset. In this figure, we can see that, the proposed FL reduces the convergence time by up to $29\%$ and $63\%$ compared to model free RL method and the standard FL method, respectively. The $29\%$ gain stems from the fact that the proposed model based RL approach can mathematically derive the transition probability across different actions and states thus speeding up the FL training process and achieving a better identification accuracy. The $63\%$ gain stems from the fact that the proposed model based RL approach enables the devices to quantize its local FL model with optimal bitwidth.

\begin{figure}[t]
	\centering
	\setlength{\belowcaptionskip}{-0.45cm}
	\includegraphics[width=12cm]{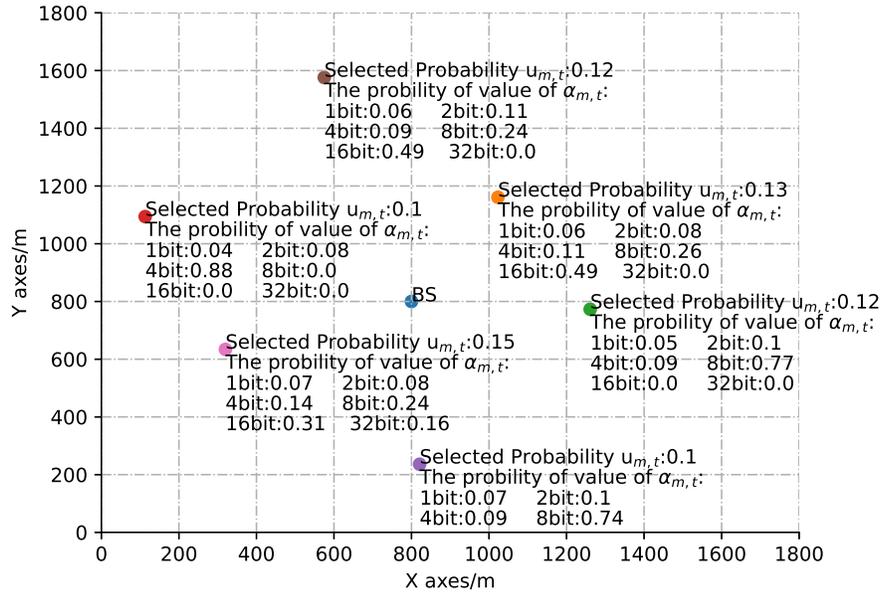}\vspace{-0.35cm}
	\centering
	\caption{Example of the optimized quantization and device selection scheme in the proposed FL framework.}
	\label{example}
 \vspace{-0.6cm}
\end{figure}

Fig. \ref{example} shows an example of the optimized device selection and quantization scheme. This simulation employs the MNIST dataset, with i.i.d. data distribution across devices. In this figure, we can see that, as the distance between the server and each device $m$ increases, the probability of quantizing the FL model on device $m$ with a high-precision bitwidth decreases. This is consistent with the fact that as the distance between the server and device $m$ increases, the time used to transmit the FL model increases under a fixed transmit power. Hence, when the data distribution across devices is i.i.d., the server will tend to have further away devices quantize their local models in lower precision so as to reduce the time used for model exchanging and training to satisfy the delay constraint $\Gamma$. Similarly, we can also see that, as the distance decreases, the probability for choosing device $m$ to participate in FL training increases. Closer devices tend to transmit higher precision FL models (when the data distributions are i.i.d.), which enables the server and the devices to obtain a FL model with a better performance of identification accuracy.

\section{Conclusion}
In this article, we developed a novel quantized FL framework in which distributed wireless devices train and transmit their locally trained FL models to a coordinating server based on variable bitwidths. We formulated an optimization problem that jointly considers the device selection and quantization scheme to minimize FL training loss while accounting for communication and computation heterogeneity across the devices. To solve this problem, we first analytically derived  the expected training convergence rate of our quantized FL framework. Our analysis showed how the expected improvement of FL training loss between two adjacent iterations depends on the device selection scheme,
the quantization scheme, and inherent properties of the model being trained. To find the tightest bound, we introduced a linear regression method for estimating these model properties according to observable training information at the server. Given these estimates, the improvement of FL performance at adjacent iterations was described as an MDP. We then proposed a model-based RL method to learn the relationship between FL performance and the choice of device selection and quantization scheme so as to converge on the policy minimizing FL loss. Numerical evaluation on real-world machine learning tasks demonstrated that the proposed methodology yields significant gains in classification accuracy and convergence speed compared to conventional approaches.

\section{Appendix}
\subsection{Proof of Lemma 1}
To prove Lemma 1, we first rewrite $F\left(\bm g_{t+1}\left(\bm u_{t+1}\right),\alpha_{t+1}\right)$ using the second-order Taylor expansion and the $L$-smoothness of property in Assumption 1, which can be expressed as
\begin{equation}\label{eq:A1}
\begin{aligned}
\!F\left(\bm g_{t+1}\right) \leq  F\left(\hat {\bm g}_{t+1}\right)+\left( \bm g_{t+1}-\hat {\bm g}_{t+1} \right) \nabla \widetilde{F}\left(\hat {\bm g}_{t+1}\right)+\frac{L}{2}\left\|\bm g_{t+1} - \hat {\bm g}_{t+1}\right\|^2,
\end{aligned}
\end{equation}
where ${\bm g}_{t}$ and $\hat {\bm g}_{t}$ are short for ${\bm g}_{t}\left(\bm u_t,\alpha_t\right)$ and $\hat {\bm g}_{t}\left(\bm u_t,\alpha_t\right)$, respectively.

Taking expectations of both sides of (\ref{eq:A1}), we have
\begin{equation}\label{eq:A2}
\begin{aligned}
\!\mathbb{E}\left(F\left(\bm g_{t+1}\right)\right) &\leq \mathbb{E}\left(F\left(\hat {\bm g}_{t+1}\right)+\left( \bm g_{t+1}-\hat {\bm g}_{t+1} \right)\nabla \widetilde{F}\left(\hat {\bm g}_{t+1}\right)+\frac{L}{2}||\bm g_{t+1} - \hat {\bm g}_{t+1}||^2  \right)\\
&\overset{(a)}{=}\mathbb{E}\left( F\left(\hat {\bm g}_{t+1}\right)\right)+\frac{L}{2}\mathbb{E}\left(\left\|\bm g_{t+1} - \hat {\bm g}_{t+1}\right\|^2\right),
\end{aligned}
\end{equation}
where (a) stems from the fact that the unbiased quantization function $Q\left({\bm g}_{t+1},\bm \alpha\right)$ satisfies $\mathbbm{E}[\hat {\bm g}_{t+1}]= {\bm g}_{t+1}$. Similarly, we rewrite $F\left(\hat {\bm g}_{t+1}\right)$ as
\begin{equation}\label{eq:A3}
\begin{aligned}
\!F\left(\hat {\bm g}_{t+1}\right) \leq F\left({\bm g}_{t}\right)+\left( \hat {\bm g}_{t+1}- {\bm g}_{t} \right)\nabla \widetilde{F}\left({\bm g}_{t}\right)+\frac{L}{2}\left\|\hat {\bm g}_{t+1}- {\bm g}_{t}\right\|^2.
\end{aligned}
\end{equation}

Taking the expectation over (\ref{eq:A3}) and combining it with (\ref{eq:A2}), we have
\begin{equation}\label{eq:L1}
\begin{split}
\mathbb{E}\left(F\left( {\bm g}_{t+1}\right)\right)&-\mathbb{E}\left(F\left({\bm g}_{t}\right)\right) \\
&\leq   \mathbb{E}\left(\left(\hat {\bm g}_{t+1}- {\bm g}_{t}\right) \nabla \widetilde{F}\left({\bm g}_{t}\right) \right)+\frac{L}{2}\mathbb{E}\left(\left\|\hat {\bm g}_{t+1}- {\bm g}_{t}\right\|^2\right)+\frac{L}{2}\mathbb{E}\left(||\hat {\bm g}_{t+1}- {\bm g}_{t+1}||^2\right).
\end{split}
\end{equation}

Then, we substitute (\ref{eq:Noiid2}) into (\ref{eq:L1}) and have
\begin{equation}\label{eq:Lnon}
\begin{split}
\mathbb{E}\left(F({\bm g}_{t+1})\right)&-\mathbb{E}\left(F({\bm g}_{t})\right)\\
&\leq  \mathbb{E}\left( \left(\hat {\bm g}_{t+1}- {\bm g}_{t}\right) \left(\nabla F({\bm g}_{t})-\epsilon \right) \right)+\frac{L}{2}\mathbb{E}\left(||\hat {\bm g}_{t+1}-{\bm g}_{t}||^2\right)+\frac{L}{2}\mathbb{E}\left(||\hat {\bm g}_{t+1}- {\bm g}_{t+1}||^2\right).
\end{split}
\end{equation}

This completes the proof.

\subsection{Proof of Theorem 1}
To prove Theorem 1, we first investigate the gap between the expectation of the quantized model $\hat {\bm g}_{t+1}$ and the expectation of the full-precision model ${\bm g}_{t}$ in Lemma 1, which is given by
\begin{equation}\label{eq:T1}
\begin{split}
\mathbb{E}\left( \hat {\bm g}_{t+1}- {\bm g}_{t}\right)&=\mathbb{E}\left(\hat {\bm g}_{t+1}-{\bm g}_{t+1}+{\bm g}_{t+1}- {\bm g}_{t}\right)\\
&=\mathbb{E}\left(\Delta\left(\alpha_t\right)-\lambda\left(\nabla F\left({\bm g}_{t}\right)-\bm e_t\right)\right),
\end{split}
\end{equation}
where $\Delta\left(\alpha_t\right)=||\hat {\bm g}_{t+1}-{\bm g}_{t+1}||$ is the quantization error at iteration $t+1$ and $\bm e_t$ is a gradient deviation caused by the quantization errors of the local FL models that are transmitted by selected devices and the devices that do not transmit their local FL models to the server at iteration $t$. In particular, $\bm e_t$ can be expressed as
\begin{equation}\label{eq:L2}
\begin{split}
\bm e_t=\nabla F\left({\bm g}_{t}\right)-\frac{\sum\limits_{m=1}^M\sum\limits_{n\in \mathcal{N}_{m,t}} u_{m,t} \left( \nabla f\left(\hat {\bm g}_{t},\bm x_{mn},\bm y_{mn}\right) + \epsilon_m \right) }{N\!\sum\limits_{m=1}^M \!N_{m,t} u_{m,t}}.
\end{split}
\end{equation}
Substituting (\ref{eq:T1}) into Lemma 1, we have
\begin{equation}\label{eq:T2}
\begin{split}
\mathbb{E}\left(F\left({\bm g}_{t+1}\right)\right)-&\mathbb{E}\left(F\left({\bm g}_{t}\right)\right) \\
\leq & \mathbb{E}\left( \left(\hat {\bm g}_{t+1}-{\bm g}_{t}\right) \left( \nabla F\left({\bm g}_{t} \right) - \epsilon \right)\right)+\frac{L}{2}\mathbb{E}\left(||\hat {\bm g}_{t+1}-{\bm g}_{t}||^2\right)+\frac{L}{2}\mathbb{E}\left(||\hat {\bm g}_{t+1}- {\bm g}_{t+1}||^2\right)\\
=&\mathbb{E}\left( \left(\Delta\left(\alpha_t\right)-\lambda\left(\nabla F\left({\bm g}_{t}\right)-\bm e_t\right)\right) \left( \nabla F\left({\bm g}_{t} \right) - \epsilon \right) \right)\\
&+\frac{L}{2}\mathbb{E}\left(||\Delta\left(\alpha_t \right)-\lambda\left(\nabla F\left({\bm g}_{t}\right)-\bm e_t\right)||^2\right)+\frac{L}{2}\mathbb{E}\left(\left\| \Delta\left(\alpha_t \right)\right\|^2\right)\\
=&-\frac{1}{2L}||\nabla F\left({\bm g}_{t}\right)||^2+\frac{1}{2L}\mathbb{E}\left(\left\|\bm e_t+L\Delta\left(\alpha_t\right)\right\|^2\right)+\frac{L}{2}\mathbb{E}\left( \Delta\left(\alpha_t\right)^2\right)\\
\overset{(a)}{=}&-\frac{1}{2L}\left\|\nabla F\left({\bm g}_{t}\right)\right\|^2+\frac{1}{2L}\mathbb{E}\left(\left\| \bm e_t\right\|^2\right)+\mathbb{E}\left(\Delta\left(\alpha_t\right)^2\right),
\end{split}
\end{equation}
where (a) stems from the fact that $\mathbbm{E}\left(\hat {\bm g}_{t+1}\right)= {\bm g}_{t+1}$. Next, we derive $\mathbb{E}\left(\left\|\bm e_t\right\|^2\right)$, which can be given as follows
\begin{align}\label{eq:T3}
 \mathbb{E}\left(\left\|\bm e_t\right\|^2\right) 
=&\mathbb{E}\left(\left\|\nabla F\left({\bm g}_{t}\right)-\frac{\sum\limits_{m=1}^M\sum\limits_{n\in \mathcal{N}_{m,t}}u_{m,t}\nabla f\left(\hat {\bm g}_{t}\right) }{N\sum\limits_{m=1}^M N_{m,t} u_{m,t}}\right\| ^2\right) \notag \\
=&\mathbb{E}\left(\left\|\nabla F\left({\bm g}_{t}\right)-\frac{\sum\limits_{m=1}^M\sum\limits_{n\in \mathcal{N}_{m,t}}u_{m,t} \left( \nabla f\left({\bm g}_{t}\!+\!\Delta\left(\alpha_t\right)\right) + \epsilon_m \right) }{\sum\limits_{m=1}^M N_{m,t} u_{m,t}}\right\| ^2 \right) \notag \\ 
\overset{(a)}{\leq}&\mathbb{E}\left(\left\|\nabla F\left({\bm g}_{t}\right)-\frac{\sum\limits_{m=1}^M\sum\limits_{n\in \mathcal{N}_{m,t}}u_m\nabla \left[f\left({\bm g}_{t}\right)+\Delta\left(\alpha_t\right)\nabla f\left({\bm g}_{t}\right)+ \epsilon_m+\frac{1}{2}||\bm o||^2 \right]}{\sum\limits_{m=1}^M \!N_{m,t} u_{m,t}}\right\| ^2 \right) \notag \\
\overset{(b)}{\leq}&\mathbb{E}\left(\left\|\nabla F\left({\bm g}_{t}\right)-\frac{\sum\limits_{m=1}^M\sum\limits_{n\in \mathcal{N}_{m,t}}u_{m,t}\left(\nabla f\left({\bm g}_{t}\right)+ \epsilon_m +\Delta\left(\alpha_t\right)L \right)}{\sum\limits_{m=1}^M \!N_{m,t} u_{m,t}}\right\| ^2 \right) \\ \notag
\end{align}
where $f\left({\bm g}_{t}\right)$ is short for $f\left({\bm g}_{t},\bm x_{mn},\bm y_{mn}\right)$, (a) stems from the second-order Taylor expansion of $f\left({\bm g}_{t}+\Delta\left(\alpha_t\right),\bm x_{mn},\bm y_{mn}\right)$, (b) stems from the twice-continuously differentiable of $f\left({\bm g}_{t},\bm x_{mn},\bm y_{mn}\right)$ with $\mu {\bm I} \! \preceq\! \nabla^2 f\left({\bm g}_{t},\bm x_{mn},\bm y_{mn}\right)\! \preceq\! L {\bm I}$. The inequality equation in (\ref{eq:T3}) is achieved by the triangle-inequality (i.e., $\left\| \nabla\! f({\bm g}_{t},\bm x_{mn},\bm y_{mn}) \right\| \leq \sqrt{\zeta_1\!+\!\zeta_2\left\| \nabla F({\bm g}_{t})\right\|^2}$). Substituting the triangle-inequality and $\sum\limits_{m=1}^M N_{m,t} u_{m,t}=A$ into (\ref{eq:T3}), we have
\begin{align}\label{eq:T44} \notag
\mathbb{E}\left(||\bm e_t||^2\right)&=\mathbb{E}\left(\left\| \frac{\left(N-A\right)\sum\limits_{m=1}^M\sum\limits_{n\in \mathcal{N}_{m,t}}\nabla \!f({\bm g}_{t})+ \epsilon_m}{NA}-\frac{1}{A}\sum\limits_{m=1}^M u_{m,t} L\Delta(\alpha_t) \right.\right. \\ \notag
&\left. \left.\qquad
+\frac{1}{N}\sum\limits_{m=1}^M N_m (1-u_{m,t})\sum\limits_{n\in \mathcal{N}_{m,t}}\nabla f({\bm g}_{t})+ \epsilon_m \right\|\right) ^{2}\\ \notag
 & \leq \mathbb{E}\left( \frac{1}{NA}\left(N-A\right)\sum\limits_{m=1}^M\sum\limits_{n\in \mathcal{N}_{m,t}}\left\|\nabla \!f({\bm g}_{t})+ \epsilon_m \right\|+\frac{1}{A}\sum\limits_{m=1}^M u_{m,t} L  \left\|\Delta(\alpha_t)\right\| \right. \\ \notag
&\left. \qquad +\frac{1}{N}\sum\limits_{m=1}^M N_m (1-u_{m,t})\!\sum\limits_{n\in \mathcal{N}_{m,t}}\left\|\nabla f({\bm g}_{t})+ \epsilon_m\right\|\right) ^{2}\\ \notag
&\quad\!\! \overset{(a)}{\leq}\mathbb{E}\left( \!\frac{\!\left(N-A\right)\!\sum\limits_{m=1}^M\sum\limits_{n\in \mathcal{N}_{m,t}}\left\|\nabla f({\bm g}_{t}) + \epsilon_m\right\|}{N}+ L  \left\|\Delta(\alpha_t)\right\|+\frac{\left(N\!\!-\!\!A\right)\!\sum\limits_{m=1}^M\sum\limits_{n\in \mathcal{N}_{m,t}}\!\!\left\|\nabla \!f({\bm g}_{t})+ \epsilon_m\right\|}{N}\!\right) ^{2}\\ \notag
&\quad\!\! {=}\mathbb{E}\left(  \!\frac{2\left(N-A\right)\!\sum\limits_{m=1}^M\sum\limits_{n\in \mathcal{N}_{m,t}}\!\!\left\|\nabla f({\bm g}_{t})+ \epsilon_m \right\|}{N}+ L  \left\|\Delta(\alpha_t)\right\|\right)^{2}\\ \notag
&\quad\!\!\overset{(b)}{\leq}\left(\mathbb{E}\left\|\Delta(\alpha_t)\right\|+1\right)\left(\frac{4\left(N-A\right)^2\left(\zeta_1 + \zeta_2\left\| \nabla F({\bm g}_{t})\right\|^2 + B\epsilon^2 \right)}{N^2}+L^2\mathbb{E}\left\|\Delta(\alpha_t)\right\|  \right) \\
\end{align}
where (a) stems from the fact that  $U\! \leq \!A$ and (b) stems from the inequality $2\bm a \bm b \!\leq\! \left\|\Delta(\alpha_t)\right\| \bm a^2\!+\! \left\|\Delta(\alpha_t)\right\|^{-1} \bm b^2$ for any two vectors $\bm a$ and $\bm b$ with scalar  $\left\|\Delta(\alpha_t)\right\| \geq 0$. Substituting (\ref{eq:T44}) into (\ref{eq:T2}), we have
\begin{equation}\label{eq:L4}
\begin{split}
\mathbb{E}\left(F( {\bm g}_{t+1})\right)&-\mathbb{E}\left(F\left({\bm g}_{t}\right)\right)\\
&\leq-\frac{1}{2L}||\nabla F\left({\bm g}_{t}\right)||^2+\frac{1}{2L}\mathbb{E}\!\left(||\bm e_t||^2\right)+\mathbb{E}\left(\Delta\left(\alpha_t\right)^2\right)\\
&\leq-\frac{1}{2L}||\nabla F({\bm g}_{t})||^2+\mathbb{E}\left(\Delta\left(\alpha_t\right)^2\right)\\
&  \quad \!\! +\frac{1}{2L}\mathbb{E}\left(\left\|\Delta(\alpha_t)\right\|+1\right)\left(\frac{4\left(N-A\right)^2\! \left(\zeta_1+\zeta_2\left\| \nabla F({\bm g}_{t})\right\|^2 + B\epsilon^2\right)}{N^2}+ L^2 \mathbb{E}\left\|\Delta\left(\alpha_t\right)\right\|\right)\\
&\hspace{-1mm}=\frac{1}{2L}\left(-1+\frac{4\left(N-A\right)^2\left(\mathbb{E}\left\|\Delta\left(\alpha_t\right)\right\|+1\right)\zeta_2}{N^2}\right)||\nabla F({\bm g}_{t})||^2\\
&\quad \!\! +\frac{\mathbb{E}\left\|\Delta\left(\alpha_t\right)\right\|+1}{2L}\left(\frac{4\left(N-A\right)^2\left(\zeta_1+B\epsilon^2\right)}{N^2} +L^{2} \mathbb{E}\left\|\Delta\left(\alpha_t\right)\right\| \right)+\mathbb{E}\left(\Delta\left(\alpha_t\right)^2\right).
\end{split}
\end{equation}

\noindent This completes the proof.

\subsection{Proof of Theorem 2}

\begin{IEEEproof} According to [Lemma 4.4, \cite{ZAZ}], the gap between the gradients aggregated by the server and the gradients trained by each device is $\sum_{i=1}^M ||\nabla F({\bm g}_{t})- \sum\limits_{n\in \mathcal{N}_{m,t}}\nabla f (\bm g_{t},\bm x_{mn},\bm y_{mn})||=M\Upsilon^2$ where $\Upsilon$ depends on $C$ and $\iota$. In our setting, the gradient deviation caused by the quantization errors of the local FL models that are trained by selected devices using nonconvex loss functions and the devices that do not transmit their local FL models to the server at iteration $t$ can be expressed as
\begin{align}\label{eq:RT15}
\mathbb{E}||\bm e_t||^2&=\mathbb{E}\left( \left\|\sum\limits_{m=1}^M\nabla F\left({\bm g}_{t}\right)- \sum\limits_{m=1}^M\sum\limits_{n\in \mathcal{N}_{m,t}}\nabla f\left( {\bm g}_{t}\right)+ \sum\limits_{m=1}^M\sum\limits_{n\in \mathcal{N}_{m,t}}\nabla f\left( {\bm g}_{t}\right)-\frac{\sum\limits_{m=1}^M\sum\limits_{n\in \mathcal{N}_{m,t}} u_{m,t}\nabla f\left(\hat {\bm g}_{t}\right) }{N\!\sum\limits_{m=1}^M \!N_{m,t} u_{m,t}} \right\|^2 \right) \notag\\
&\leqslant \mathbb{E} \left( \left\| \sum\limits_{m=1}^M \sum\limits_{n\in \mathcal{N}_{m,t}}\nabla f\left( {\bm g}_{t}\right)-\frac{\sum\limits_{m=1}^M\sum\limits_{n\in \mathcal{N}_{m,t}} u_{m,t}\nabla f\left(\hat {\bm g}_{t}\right) }{N\!\sum\limits_{m=1}^M \!N_{m,t} u_{m,t}} \right\|^2 \right) +M^2\Upsilon^4 \notag\\
&\overset{(a)}{\leq}\mathbb{E}\left(\left\|\nabla F\left({\bm g}_{t}\right)-\frac{\sum\limits_{m=1}^M\sum\limits_{n\in \mathcal{N}_{m,t}}u_{m,t} \left( \nabla f\left({\bm g}_{t}\!+\!\Delta\left(\alpha_t\right)\right) + \epsilon_m \right) }{\sum\limits_{m=1}^M N_{m,t} u_{m,t}}\right\| ^2 \right) + M^2\Upsilon^4,
\end{align}
where (a) stems from Condition 2 and [Lemma 4.4, \cite{ZAZ}]. Then, using a similar argument on how we obtain (41), we have
\begin{equation}\label{eq:RT16}
\begin{aligned}
||\bm e_t||^2\overset{}{\leq}\left(\mathbb{E}\left\|\Delta(\alpha_t)\right\|+1\right)\left(\frac{4\left(N-A\right)^2\left(\zeta_1 + \zeta_2\left\| \nabla F({\bm g}_{t})\right\|^2 + B\epsilon^2 \right)}{N^2}+L^2\mathbb{E}\left\|\Delta(\alpha_t)\right\|  \right) +M^2\Upsilon^4 .
\end{aligned}
\end{equation}
Finally, Theorem 1 can be rewritten as
\begin{equation}\label{eq:T5}
\begin{split}
\mathbb{E}\left(F( {\bm g}_{t+1})\right)&-\mathbb{E}\left(F\left({\bm g}_{t}\right)\right)\\
&\leqslant\frac{1}{2L_{\gamma}}\left(-1+\frac{4\left(N-A\right)^2\left(\mathbb{E}\left\|\Delta\left(\alpha_t\right)\right\|+1\right)\zeta_2}{N^2}\right)||\nabla F({\bm g}_{t})||^2 + \frac{M^2\Upsilon^4}{2L_{\gamma}}\\
&\quad \!\! +\frac{\mathbb{E}\left\|\Delta\left(\alpha_t\right)\right\|+1}{2L_{\gamma}}\left(\frac{4\left(N-A\right)^2\zeta_1}{N^2} +L^{2} \mathbb{E}\left\|\Delta\left(\alpha_t\right)\right\| \right)+\mathbb{E}\left(\Delta\left(\alpha_t\right)^2\right).
\end{split}
\end{equation}
This completes the proof.
\end{IEEEproof}

\bibliographystyle{IEEEbib}
\renewcommand{\baselinestretch}{1.4}
\bibliography{BNN}
\end{document}